\documentclass[10pt,journal,compsoc]{IEEEtran}

\ifCLASSOPTIONcompsoc
  \usepackage[nocompress]{cite}
\else
  \usepackage{cite}
\fi

\ifCLASSINFOpdf
\else
\fi

\usepackage{amsmath,amsfonts}
\usepackage{algorithmic}
\usepackage{algorithm}
\usepackage{array}
\usepackage[caption=false,font=normalsize,labelfont=sf,textfont=sf]{subfig}
\usepackage{textcomp}
\usepackage{stfloats}
\usepackage{url}
\usepackage{verbatim}
\usepackage{graphicx}
\usepackage{threeparttable} 
\usepackage{booktabs} 
\usepackage{cite}
\usepackage{multirow}
\usepackage{overpic}

\usepackage{color}

\def\ie{\emph{i.e.}}

\newcommand{\ebf}[1]{\boldsymbol{#1}}

\definecolor{dt}{gray}{0.50}

\begin{document}
%
\title{Fusion from Decomposition: A Self-Supervised Approach for Image Fusion and Beyond}

\author{Pengwei Liang, Junjun Jiang,~\IEEEmembership{Senior Member,~IEEE,} Qing Ma, Xianming Liu,~\IEEEmembership{Member,~IEEE,} and Jiayi Ma,~\IEEEmembership{Senior Member,~IEEE}

\thanks{P. Liang, J. Jiang, Q. Ma and X. Liu are with the School of Computer Science and Technology, Harbin Institute of Technology, Harbin 150001, China. E-mail: \{erfect2020, qingma2016\}@gmail.com, \{jiangjunjun, csxm\}@hit.edu.cn.}
\thanks{J. Ma is with the Electronic Information School, Wuhan University, Wuhan 430072, China. E-mail: jyma2010@gmail.com.}
}


%
%

\markboth{Journal of \LaTeX\ Class Files,~Vol.~14, No.~8, August~2015}%
{Shell \MakeLowercase{\textit{et al.}}: Bare Advanced Demo of IEEEtran.cls for IEEE Computer Society Journals}
%


\IEEEtitleabstractindextext{%
\begin{abstract}
Image fusion is famous as an alternative solution to generate one high-quality image from multiple images in addition to image restoration from a single degraded image. The essence of image fusion is to integrate complementary information from source images. Existing fusion methods struggle with generalization across various tasks and often require labor-intensive designs, in which it is difficult to identify and extract useful information from source images due to the diverse requirements of each fusion task. Additionally, these methods develop highly specialized features for different downstream applications, hindering the adaptation to new and diverse downstream tasks. To address these limitations, we introduce DeFusion++, a novel framework that leverages self-supervised learning (SSL) to enhance the versatility of feature representation for different image fusion tasks. DeFusion++ captures the image fusion task-friendly representations from large-scale data in a self-supervised way, overcoming the constraints of limited fusion datasets. Specifically, we introduce two innovative pretext tasks: common and unique decomposition (CUD) and masked feature modeling (MFM). CUD decomposes source images into abstract common and unique components, while MFM refines these components into robust fused features. Jointly training of these tasks enables DeFusion++ to produce adaptable representations that can effectively extract useful information from various source images, regardless of the fusion task. The resulting fused representations are also highly adaptable for a wide range of downstream tasks, including image segmentation and object detection. DeFusion++ stands out by producing versatile fused representations that can enhance both the quality of image fusion and the effectiveness of downstream high-level vision tasks, simplifying the process with the elegant fusion framework. We evaluate our approach across three publicly available fusion tasks involving infrared and visible fusion, multi-focus fusion, and multi-exposure fusion, as well as on downstream tasks. The results, both qualitative and quantitative, highlight the versatility and effectiveness of DeFusion++.

\end{abstract}

}

\maketitle

\IEEEdisplaynontitleabstractindextext

%
\IEEEpeerreviewmaketitle

\ifCLASSOPTIONcompsoc
\IEEEraisesectionheading{\section{Introduction}\label{sec:introduction}}
\else
\section{Introduction}
\label{sec:introduction}
\fi

%
%
%
%

\IEEEPARstart{I}{mage} fusion is a fundamental technique in image processing and computer vision that amalgamates multiple images to generate a single fused image, retaining the most desirable features from the contributing inputs. The resulting image typically exhibits enhanced quality, reduced uncertainty, and a more detailed representation of the scene compared to the individual source image \cite{zhang2021image}. Over the years, the significance of image fusion has increased dramatically, driven by its applications in diverse fields such as remote sensing \cite{ma2024reciprocal}, medical imaging \cite{xu2021emfusion}, surveillance \cite{liu2011objective}, and virtual reality \cite{wang2021deep}. 
In recent years, significant advancements have been made in image fusion, attracting considerable attention \cite{ma2019infrared, ma2020infrared, xu2020u2fusion, ma2022swinfusion}. Pioneering works based on deep learning are dedicated to addressing specific fusion tasks such as infrared and visible image fusion (IVF) \cite{ma2019infrared, li2018densefuse}, medical image fusion \cite{xu2021emfusion}, multi-focus image fusion (MFF) \cite{liu2017multi}, and multi-exposure image fusion (MEF) \cite{ram2017deepfuse, liu2024coconet}. Designing task-specific fusion methods can be labor-intensive. Unified fusion methods address various fusion problems within a single unified model, enabling these tasks to promote one another \cite{xu2020u2fusion}.
The unified fusion methods not only yield aesthetically pleasing results, but also demonstrate potential for generalizing across different fusion tasks.

\begin{figure*}[htbp]
    \centering
    \includegraphics[width=18cm]{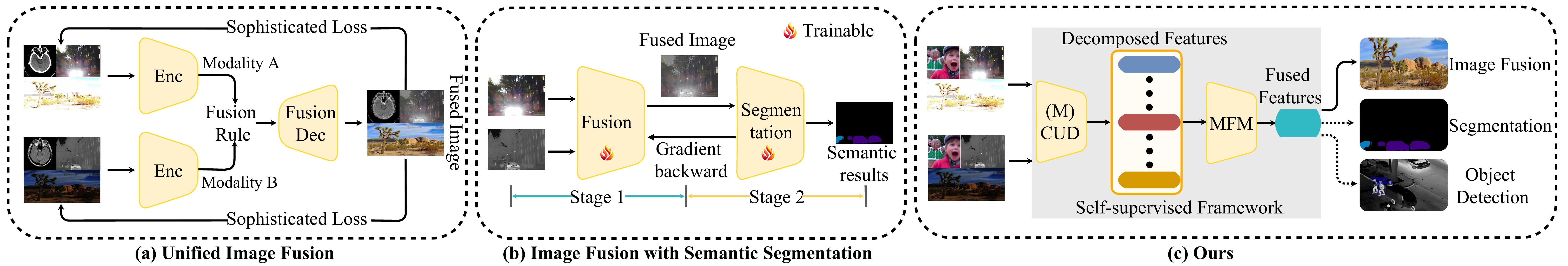}
    \caption{ Overview of traditional image fusion approaches illustrated in (a) and (b), respectively. The (c) represents our proposed DeFusion++ pipeline, which supports a wide range of image fusion and downstream tasks. In our method, we propose two self-supervised pretext task: multi-modal common and unique decomposition (CUD) and masked feature modeling (MFM).
    }
    \label{fig: introduction}
\end{figure*}

Nevertheless, these unified models face significant challenges in image fusion, particularly in identifying and refining the most useful information from multiple source images into a single fused image. The primary objective of image fusion is to integrate complementary information from various source images. However, the nature of this complementary information varies significantly across different fusion tasks. For example, in IVF, the pixel intensity from the infrared image and the gradient information from the visible image are particularly valuable \cite{ma2019fusiongan}. In MFF, the focus regions are essential, whereas in MEF, preserving rich texture information is crucial. To address these diverse requirements, some unified fusion models employ sophisticated loss functions and tailored fusion rules to harmonize the characteristics of the source images \cite{zhang2020rethinking}, as illustrated in Fig. \ref{fig: introduction}(a). However, with the advancement of imaging devices, designing such tailored loss functions and fusion rules has become increasingly challenging. One notable example is that the latest high-quality infrared images may also exhibit rich textures, which contradicts our previous knowledge. Therefore, there is a pressing need for a straightforward and effective method to extract useful information from source images to overcome these challenges.

In addition, the task of image fusion is born with two inherent goals. The first is for visual perception. Specifically, in the context of MEF, low dynamic range images often fail to capture the full spectrum of scene information due to underexposure and overexposure. Fusing those low dynamic range images into a high dynamic range image substantially enhances visual quality, making this technique essential for applications like high-quality film production.
The second is to support downstream vision tasks such as object detection and image segmentation. For example, integrating visible and infrared images enhances the accuracy of segmentation under varying lighting conditions \cite{zhao2023cddfuse, liu2022target}. In such scenarios, the primary objective is to generate a segmentation map, with the fused image serving merely as an auxiliary output. However, simultaneously considering these two aspects often receive insufficient attention from conventional fusion methods. To bridge this gap, recent works in IVF have incorporated loss functions tailored to specific downstream tasks into the fusion algorithm \cite{tang2022image, liu2022target}, as illustrated in Fig. \ref{fig: introduction}(b). In these works, the fusion process is structured into two distinct stages. Initially, the network is optimized specifically for image fusion. Subsequently, it shifts focus to fine-tune the network for downstream tasks. They expect these stages to support and enhance each other, and enable the network to not only generate aesthetically superior fused images but also to improve performance on specific high-level vision tasks. However, since these methods are custom-designed for specific tasks, the learned features may become too specialized to adapt to other types of tasks.

To address these challenges, we introduce a pioneering image fusion framework that harnesses the power of self-supervised learning (SSL). Our framework explores the feature representation capability from large-scale data and is not limited by the number of paired datasets in image fusion tasks. By utilizing SSL, which focuses on learning the intrinsic structure of data, the framework can extract more generalizable features, demonstrating superior performance in cross-domain and cross-task adaptability. It introduces fusion-related pretext tasks, avoiding the need for sophisticated loss functions and tailored fusion rules for each type of fusion task. Furthermore, it produces generic fused features, allowing excellent performance across various downstream tasks. Specifically, we introduce DeFusion++, a self-supervised learning framework focused on generating powerful fused features that can be easily transformed into fused images or used to support downstream tasks. Specifically, we redefine the core concept of image fusion by asserting that source images can be effectively decomposed into unique and common components, and subsequently the target fusion image is synthesized by methodically combining these components. Based on this concept, we develop two novel pretext tasks: common and unique decomposition (CUD) and masked feature modeling (MFM), as illustrated in Fig. \ref{fig: introduction}(c). The CUD is tailored to decompose source images, including those from multi-modal sources, into their common and unique components. The MFM refines the abstract common and unique components into more useful and effective fused features. By jointly training CUD and MFM, DeFusion++ generates abstract byproducts, \ie, common and unique components, along with the target fused features. In this process, CUD circumvents the requirement for sophisticated, task-specific loss functions and manual adjustments to fusion rules. Concurrently, MFM significantly enhances the robustness and effectiveness of feature representations within the fusion network, allowing it to support a broader range of downstream vision tasks, including the generation of fused images. The characteristics and contributions of our work are summarized as follows:
\begin{itemize}
    \item We introduce DeFusion++, a self-supervised fusion method that re-examines the image fusion problem from the perspective of image decomposition. Integral to our method is a novel pretext task, common and unique decomposition (CUD). Leveraging CUD, DeFusion++ can support various image fusion tasks, circumventing the necessity for sophisticated loss functions and customized fusion rules.
    
    \item We also design a self-supervised pretext task, masked feature modeling (MFM), which refines decomposed representations into robust, fused representations. The fused representations are capable of supporting a variety of downstream tasks, including image fusion, image segmentation, and object detection.
    
    \item To the best of our knowledge, the introduced framework is the first to produce adaptable fused representations that effectively support a wide range of image fusion and downstream tasks simultaneously. It successfully addresses the requirements of both visual perception and application-specific needs. 
    
    \item We apply our unified fusion network to a diverse set of data and test it on three fusion tasks including infrared-visible fusion, multi-focus fusion, and multi-exposure fusion. Furthermore, we evaluate the proposed method on two downstream tasks of image segmentation and object detection. Qualitative and quantitative results validate the effectiveness of DeFusion++.
\end{itemize}

A preliminary version of this work appeared as DeFusion \cite{liang2022fusion}. The current version extends the original contributions across three aspects. First, we refine the CUD task and introduce multi-modal CUD (MCUD), which is better suited for handling multi-modal source images. Second, we develop a novel pretext task, MFM, aimed at enhancing the representation capabilities of fused features. Third, our enhanced self-supervised framework not only generates fused images but also produces robust fused feature representations suitable for two typical downstream tasks of image segmentation and object detection.

\section{Related Work}

\begin{figure*}[htbp]
    \centering
    \includegraphics[width=18cm]{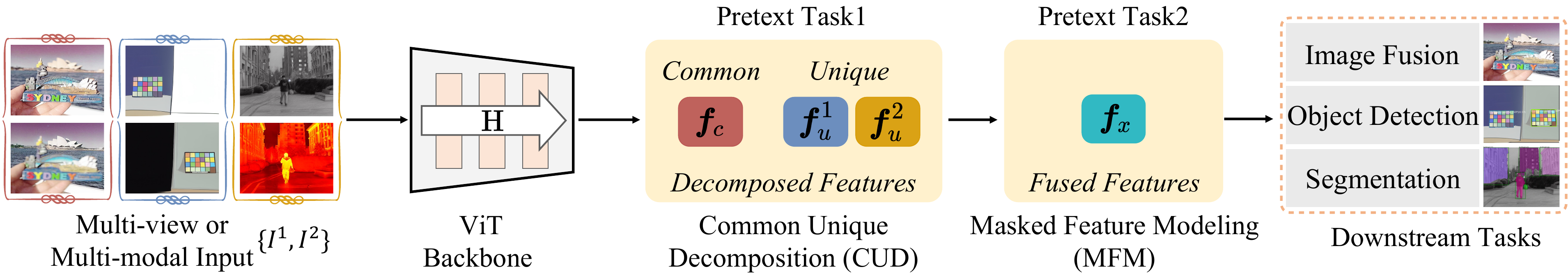}
    \caption{The overall framework of DeFusion++. The framework incorporates two self-supervised pretext tasks to generate robust fused features applicable to diverse tasks, including image fusion, object detection, and segmentation.}
    \label{fig: overall}
    \vspace{-3mm}
\end{figure*}

\subsection{Image Fusion Based on Deep Learning}

From the perspective of supervision paradigms, existing image fusion methods can be roughly categorized into three classes: supervised, unsupervised, and self-supervised. A representative model of the supervised method is IFCNN \cite{zhang2020ifcnn}, which is designed specifically for multi-focus image fusion tasks. Supervised methods require ground truth data for training; however, this is often unavailable in many image fusion tasks. To address this issue, numerous unsupervised methods have been proposed. In unsupervised methods, two main components are critical: the design of the loss function and the fusion rule. The loss function typically relies heavily on prior information about the specific fusion task, such as pixel intensity of infrared images \cite{ma2019fusiongan}, texture details of visible images \cite{ma2020infrared}, illumination \cite{tang2022piafusion}, salient information \cite{li2023lrrnet}, and gradients \cite{zhang2020rethinking, zhu2024taskcustomized}. As for the fusion rules, they mainly include two classes: manually designed \cite{ram2017deepfuse, li2018densefuse, li2020nestfuse, deng2020deep, hu2023zmff, xu2020u2fusion, shang2024holistic} and learnable \cite{xu2023unsupervised, hong2024merf, jiang2023meflut, zou2024enhancing}. Unsupervised learning requires substantial prior knowledge tailored to the specific fusion dataset. To address these challenges, self-supervised learning methods have been proposed. These methods typically employ a pretext task to obtain a powerful representation, such as pixel intensity transformation, brightness transformation \cite{qu2024trans2fuse}, and solving puzzles \cite{qu2022transmef}. However, self-supervised methods still struggle to design specific pretext tasks for each fusion task and to integrate these tasks for comprehensive performance improvement. Another drawback of these methods is their sole focus on creating a visually pleasing fused image with higher objective evaluations. Some superior methods may handle various image fusion tasks simultaneously, but they often neglect the downstream tasks of fusion.

\subsection{Joint Image Fusion and High-level Tasks}

Recent advancements have sparked considerable interest in integrating high-level vision tasks with infrared-visible image fusion \cite{tang2022image, liu2022target, tang2023rethinking}. Tang \emph{et al.} initially proposed SeAFusion \cite{tang2022image}, which combines a semantic segmentation network with an image fusion network. This design optimizes the segmentation and fusion tasks simultaneously. In a similar vein, Liu et al. introduced TarDAL \cite{liu2022target}, which replaces the semantic segmentation network in SeAFusion with an object detection network, thereby achieving leading object detection performance alongside pleasing fusion outcomes. Moving beyond serial integration, Tang \emph{et al.} proposed PSFusion \cite{tang2023rethinking}, where they explore a parallel integration of the semantic segmentation model with the image fusion network. The advantage of this parallel architecture is that it explicitly injects semantic information into the fusion network at the feature level. However, these methods exhibit two major drawbacks: (i) they are limited to the specific task of infrared-visible image fusion; (ii) the design that combines high-level vision tasks generally requires an additional network to support the training. In our work, we address these issues by employing a self-supervised learning framework. This framework enables the backbone not only to handle various fusion tasks but also to produce semantic features that support various downstream tasks involving semantic segmentation and object detection.

\subsection{Self-supervised Learning}
Self-supervised learning is a paradigm used to derive useful representations from large unlabeled datasets \cite{liu2021self, zhou2023dynamics}. This approach typically involves designing specific pretext tasks to extract powerful representations \cite{doersch2015unsupervised}. Common pretext tasks include solving jigsaw puzzles \cite{noroozi2018boosting}, recognizing image orientation \cite{komodakis2018unsupervised}, learning to count \cite{noroozi2017representation}, and image colorization \cite{zhang2016colorful}. Among these, contrastive learning and masked image modeling (MIM) are particularly popular. Contrastive learning, exemplified by methods such as SimCLR \cite{chen2020simple} and MoCo \cite{he2020momentum}, focuses on distinguishing between similar (positive) and dissimilar (negative) pairs of data points. On the other hand, a leading approach in masked image modeling is the masked autoencoder (MAE) proposed by He \emph{et al.} \cite{he2022masked}, which involves randomly masking regions of an input image and predicting the masked pixels. The MAE has garnered significant attention within the computer vision community for its impressive performance, inspiring subsequent studies that have achieved new state-of-the-art results across various fields \cite{zhou2021ibot, fang2023corrupted, ren2023tinymim}. Inspired by the success of MAE, we propose a masked feature modeling pretext task specifically tailored for image fusion tasks in this study. Our approach aims to enhance the capability of extracting representations by integrating domain-specific challenges inherent in image fusion, thereby improving the effectiveness of the fusion process.


\section{Methods}

We propose a method that facilitates image fusion, including multi-modal scenarios, by leveraging a self-supervised learning framework. The framework incorporates two pivotal pretext tasks, common and unique decomposition (CUD) and masked feature modeling (MFM), to produce a good feature representation that simultaneously supports both image-level visual fusion and downstream high-level tasks. In this section, we will elaborate on these pretext tasks and demonstrate how to effectively apply the trained model to a range of tasks, including image fusion and the associated downstream tasks such as multi-modal image segmentation and object detection.

\begin{figure*}[htbp]
    \centering
    \includegraphics[width=18cm]{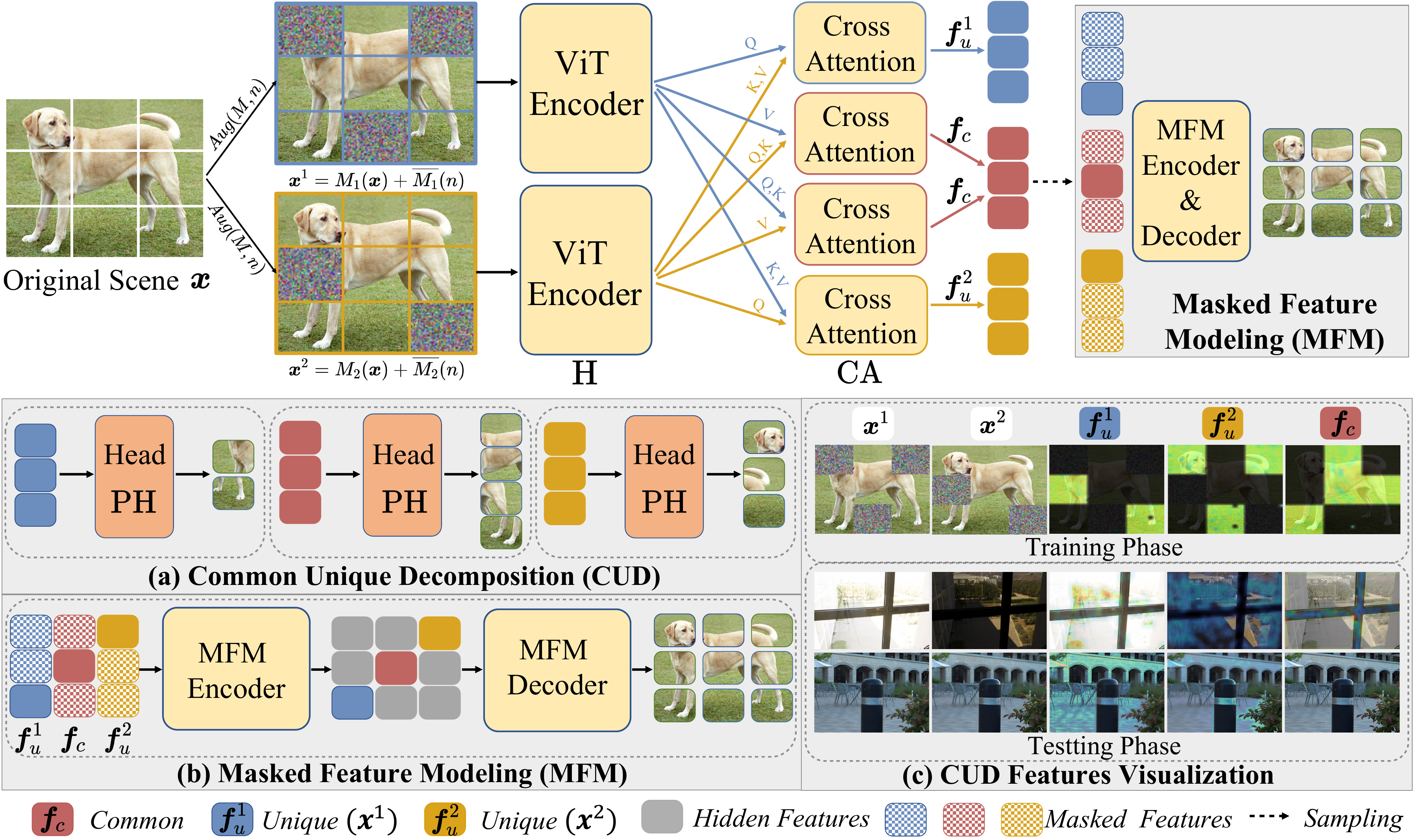}
    \caption{The paradigm of common and unique decomposition (CUD) and masked feature modeling (MFM). The (a) depicts the detailed process of CUD. The (b) illustrates the main idea of MFM. In (c), we apply heatmaps to the source images to visually highlight the areas focused on by the unique and common features, showing where these features are identified in training and testing phase.
    }
    \label{fig: cud}
\end{figure*}

\subsection{Overall Framework}

The proposed framework is designed to derive the most effective feature representation, denoted as $\ebf{f}_x$, from source images $I^1$ and $I^2$. The representation can be readily transformed into a fused image or applied for downstream tasks such as segmentation and object detection. As depicted in Fig. \ref{fig: overall}, the framework includes two principal pretext tasks: the common and unique decomposition (CUD) and the masked feature modeling (MFM). The CUD partitions the representation into three components: $\ebf{f}_u^1$ and $\ebf{f}_u^2$ (representing unique characteristics of each input), and $\ebf{f}_c$ (capturing common features across all inputs). CUD is specifically adapted for multi-modal images and multiple views through two custom-designed subtasks, \emph{i.e.}, common feature extraction and unique feature extraction. Following decomposition, the MFM processes these components ($\ebf{f}_u^1, \ebf{f}_u^2, \ebf{f}_c$) to enhance and refine the features into a more potent representation, $\ebf{f}_x$. The enriched feature is crucial for generating high-quality fused images or detailed segmentation maps.


\subsection{Common Unique Decomposition}

The proposed common unique decomposition (CUD) follows a typically self-supervised learning paradigm. Within the paradigm, the backbone network extracts robust representations that are proven to be directly applicable to downstream tasks \cite{jing2020self}. Specifically, CUD is tailored to image fusion, designed to produce representations that facilitate the efficient generation of fused images.
\subsubsection{Masked Transformation}

Following the conventional self-supervised frameworks, the CUD task leverages an unlabeled image dataset $D$ to cultivate a robust feature representation network, as depicted in Fig. \ref{fig: cud}. Each image $\ebf{x}$ within $D$, denoted as $\ebf{x} \in \mathbb{R}^{H\times W \times 3}$, is transformed through a series of random data augmentations via a predefined set of image transformations, referred to as $Aug$. These augmentations transform the original images into distorted views, $\ebf{x}^i$. An cross attention network then processes these views, tasked with capturing both common and unique feature representations essential for effective image fusion.

The proposed CUD task aims to simulate the fusion process. In this process, $\ebf{x}$ represents the original scene, and the distorted views $\{\ebf{x}^i\}_i$ simulate observations that capture one aspect of the scene. To mimic this, we introduce random masks $M_i$ and Gaussian noise $n$, collectively forming the degradation transformation $Aug$:
\begin{equation}
    \begin{aligned}
        \ebf{x}^1 = &M_1(\ebf{x}) + \Bar{M_1}(n), \\
        \ebf{x}^2 = &M_2(\ebf{x}) + \Bar{M_2}(n), \\
          s.t. \; &M_1 + M_2 \succ 0, 
    \end{aligned}
    \label{decomposition equation}
\end{equation}
where $\Bar{M_i}$ is the logical negation operator of mask $M_i$. The constraint of $M_1 + M_2 \succ 0$ guarantees that the original information of scene is comprehensively represented across the augmented views. This setup reflects the image fusion paradigm, where each source image retains unique information yet shares common elements with others. The Gaussian noise $n$ ensures the independence of unique information across different views. Furthermore, the masks $M_1$ and $M_2$ are designed to be non-orthogonal to preserve shared information in the degraded views. 

\subsubsection{Cross Attention Layer }

\begin{figure}[t]
    \centering
    \includegraphics[width=9cm]{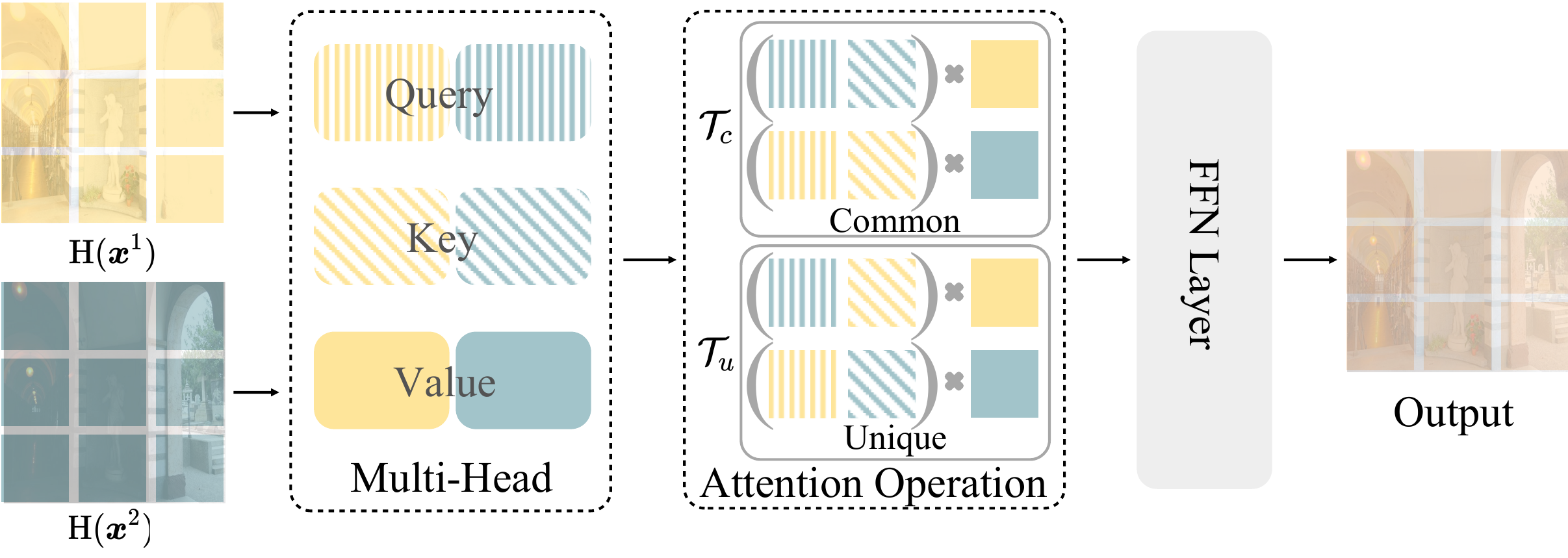}
    \caption{The mechanism of the cross attention layer, which processes two input features $\text{H}(\ebf{x}^1), \text{H}(\ebf{x}^2)$. Based on the task indicator $\mathcal{T}$, it strategically selects different $Q$ (query), $K$ (key), and $V$ (value) to specifically generate either common or unique features tailored to the demands of (M)CUD tasks.}
    \label{fig: across layer}
\end{figure}

We now introduce the cross attention layer ($\text{CA}$), designed specifically for processing multi-view/modal image inputs. The architecture of cross attention layer is pivotal for unique and common information extraction. Traditional attention mechanisms, as described in prior work \cite{vision_transformer}, utilize a single input where the attention function is defined as: $ \text{FFN}(\text{softmax}(\frac{QK^{T}}{\sqrt{d}}) V)$. Here, the query $Q$, key $K$, and value $V$ are features extracted from the input through linear transformations. The $d$ represents the dimension of the input token, with $\sqrt{d}$ serving as a scaling factor, and the $\text{FFN}$ denotes the feed-forward networks. In contrast, our cross attention layer, $\text{CA}(\text{H}(\ebf{x}^1), \text{H}(\ebf{x}^2), \mathcal{T})$, accommodates and integrates features from multi-view/modal image inputs, $\ebf{x}^1$ and $\ebf{x}^2$. The $\mathcal{T}$ serves as the indicator to distinguish common and unique decomposition tasks. The dual-input capability allows the $\text{CA}$ to efficiently integrate and synthesize both the common and unique features across views or modalities, as illustrated in Fig. \ref{fig: across layer}. 

\subsubsection{CUD Loss}

The objective of image fusion is to combine complementary information from various source images into a single cohesive synthetic image. The goal can be achieved through two main steps: decomposition and reconstruction. During decomposition, a robust backbone comprising a vision transformer ($\text{H}$) and cross attention layers are applied to extract the common and unique representations.  Moving to the reconstruction stage, the previously extracted common and unique representations are combined to reconstruct the original scene. The common and unique representations can be represented as:
\begin{equation}
    \begin{aligned}
        \ebf{f}_c, \ebf{f}_u^1, \ebf{f}_u^2 = \text{CA}(\text{H}(\ebf{x}^1),\text{H}(\ebf{x}^2), \mathcal{T}),
    \end{aligned}
\end{equation}
where $\text{H}$ represents the vision transformer encoder, $\text{CA}$ denotes cross attention layers, and $\mathcal{T}$ serves as the indicator to distinguish common and unique decomposition tasks. The terms $\ebf{f}_c, \ebf{f}_u^1, \ebf{f}_u^2$ denote the common and unique representations, respectively. Finally, the reconstruction of the fused image is modeled as: $\text{PH}(\text{H}_{MFME}(\ebf{f}_c, \ebf{f}_u^1, \ebf{f}_u^2))$. The $\text{PH}(\text{H}_{MFME}(\cdot))$ represents the network that translates the complete scene representation from the latent feature space back into the image domain. A detailed exploration of the architecture $\text{H}_{MFME}$ will be provided in Sec. \ref{sec: MFM}.

To ensure the effectiveness of the common and unique representations, we first convert these back into image space. We then compute ground truth images from the input degraded views, which serve as a reference to guide the transformation. The loss function for CUD is formulated as:
\begin{equation}
    \begin{aligned}
        L_{s\text{-}cud}(\ebf{x}^1, \ebf{x}^2) =& \| \text{PH}(\ebf{f}_c) - M_1(\ebf{x}^1) \cap M_2(\ebf{x}^2) \|_1 \\
         &+ \| \text{PH}(\ebf{f}_u^1) - M_1(\ebf{x}^1) \cap \Bar{M_2}(\ebf{x}^2) \|_1  \\
         &+ \| \text{PH}(\ebf{f}_u^2) - \Bar{M_1}(\ebf{x}^1) \cap M_2(\ebf{x}^2) \|_1 \\
        &+ \| \text{PH}(\text{H}_{MFME}(\ebf{f}_c, \ebf{f}_u^1, \ebf{f}_u^2)) - \ebf{x} \|_1 ,
    \end{aligned}
\end{equation}
where $\text{PH}$ are shallow-layer projector heads that map features back to image space. The intersection operation, represented by $\cap$, is performed pixel by pixel. This approach ensures accurate representation of overlapping areas, while regions affected by Gaussian noise or unique to one image are set to zero.

\subsection{Adapt the CUD to Multi-modal Image}
\label{sec: adaptively}

\begin{figure}[t]
    \centering
    \includegraphics[width=8.5cm]{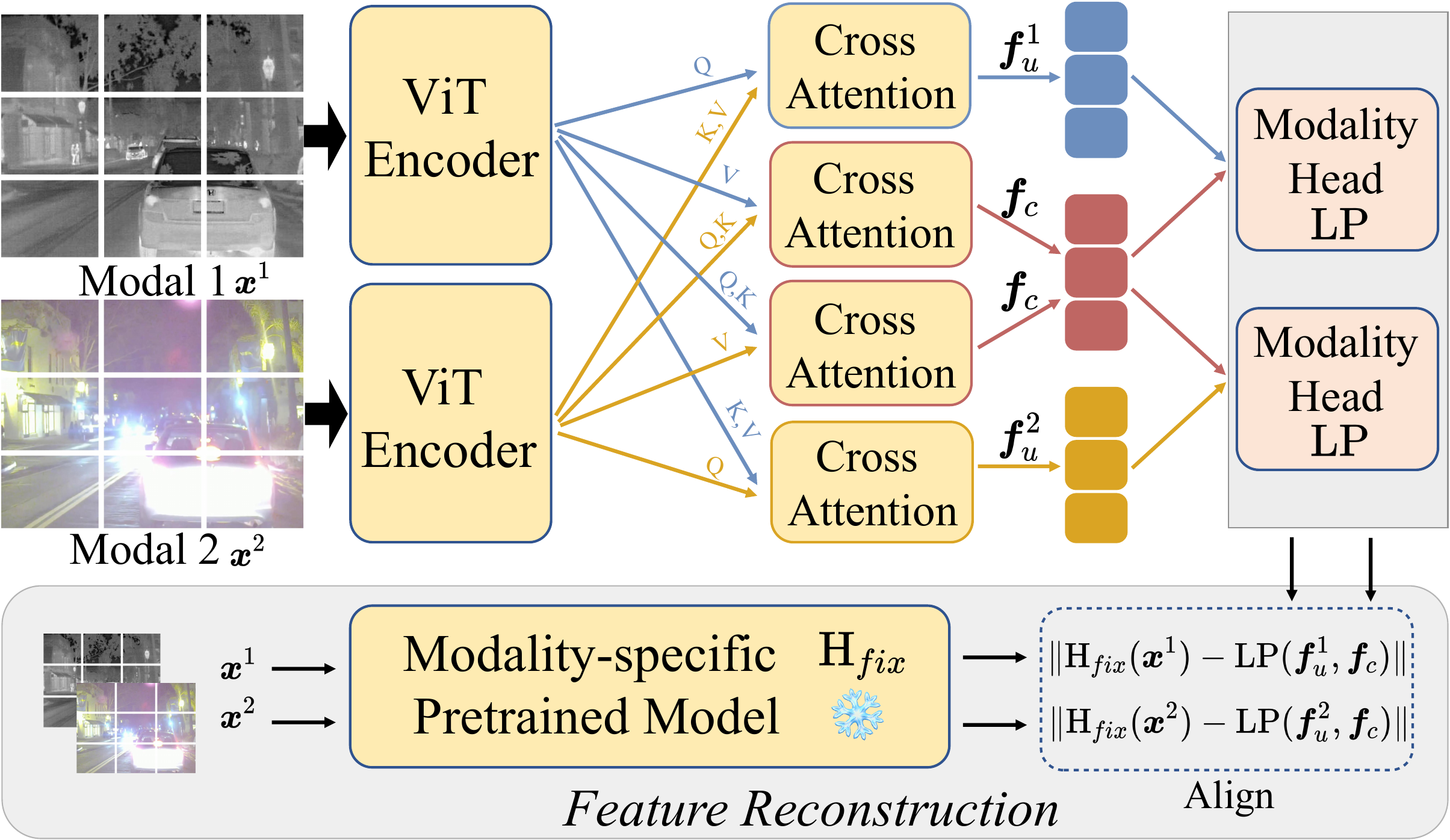}
    \caption{Architectural diagram of the  multi-modal common and unique decomposition (MCUD). By introducing the cross attention layer and the modality pretrained model, we ensure that the extracted features preserve the modality-specific information.}
    
    \label{fig: mcud}
\end{figure}

In the section above,  we introduce a preliminary pretext task, \ie, CUD, tailored specifically for image fusion. Although effective at decomposing input views into common and unique components using a backbone network, CUD primarily addresses single-modal scenes. Such a limitation restricts its ability to fully handle the complexities associated with multi-modal image fusion. Multi-modality is essential in contexts like infrared-visible image fusion, where it maximizes the extraction of useful information under diverse imaging conditions. Consider low-light conditions where infrared images can still capture vital thermal data, whereas visible images might suffer from underexposed and produce low-quality visuals. Therefore, extending the CUD framework to multi-modal CUD (MCUD) is essential. Adapting CUD to MCUD enhances the modality-specific characteristics of the extracted unique representations. It also ensures the capture of essential information from each modality, which single-modality CUD might otherwise overlook.

To achieve this goal, we propose a multi-modal common unique decomposition (MCUD) as an extension of the original CUD task to support the input of multi-modal images. The MCUD consists of two main objectives: 1) extracting common and unique information from multi-modal images; 2) ensuring the extracted unique representation retains sufficient modality-specific information. To achieve the first objective, we carefully design a cross attention layer employing an advanced attention mechanism to ensure the efficient extraction of common representations. Regarding the second objective, we introduce a pretrained multi-modal encoder alongside a novel feature alignment technique. Figure \ref{fig: mcud} depicts the detailed structure of the MCUD framework.

\subsubsection{Multi-modal Common Representations}

For the extraction of common representations, we employ a specialized configuration in our cross attention layer. In this setup, both the key $K$ and query $Q$ are derived from the same modality, while the value $V$ is sourced from the other modality. The configuration allows features from one modality to generate the attention map \ie, $\text{softmax}(\frac{QK^{T}}{\sqrt{d}})$, and features from the other modality to form the common representation based on this map. In this way, common representations are free from bias associated with modality-specific information. Consequently, in systems incorporating two modalities, the method facilitates the generation of two common representations. Each representation integrates features from both modalities, maintaining balance without favoring either one:
\begin{equation}
    \begin{aligned}
Q_i &= W^Q \cdot \text{H}(\ebf{x}^i), K_i = W^K \cdot \text{H}(\ebf{x}^i), V_i = W^V \cdot \text{H}(\ebf{x}^i), \\
    &\text{CA}(\text{H}(\ebf{x}^1), \text{H}(\ebf{x}^2), \mathcal{T}_c) = \text{FFN}(\text{softmax}(\frac{Q_2K_2^{T}}{\sqrt{d}}) V_1), \\
    &\text{CA}(\text{H}(\ebf{x}^2), \text{H}(\ebf{x}^1), \mathcal{T}_c) = \text{FFN}(\text{softmax}(\frac{Q_1K_1^{T}}{\sqrt{d}}) V_2),
\end{aligned}
\end{equation}
where $\mathcal{T}_c$ indicates the task of extracting common representations. The $W^Q, W^K, W^V$ denote the parameters of the linear layer. To ensure consistency across modalities in the common representations, we enforce a consistency constraint:
\begin{equation}
\begin{aligned}
 L&_{m\text{-}com}(\ebf{x}^1, \ebf{x}^2) = \\ & \| \text{CA}(\text{H}(\ebf{x}^1), \text{H}(\ebf{x}^2), \mathcal{T}_c) - \text{CA}(\text{H}(\ebf{x}^2), \text{H}(\ebf{x}^1), \mathcal{T}_c) \|.
\end{aligned}
\end{equation}

\subsubsection{Multi-modal Unique Representations}

For the extraction of unique representations, we employ a commonly used cross-attention architecture as described in \cite{vaswani2017attention}. In this layer, the attention maps are generated by interchanging the roles of the keys and queries from two different modalities, leading to the following formulations:
\begin{equation}
    \begin{aligned}
        \text{CA}(\text{H}(\ebf{x}^1), \text{H}(\ebf{x}^2), \mathcal{T}_u) = \text{FFN}(\text{softmax}(\frac{Q_2K_1^{T}}{\sqrt{d}}) V_1), \\
        \text{CA}(\text{H}(\ebf{x}^2), \text{H}(\ebf{x}^1), \mathcal{T}_u) = \text{FFN}(\text{softmax}(\frac{Q_1K_2^{T}}{\sqrt{d}}) V_2),
    \end{aligned}
    \label{eq: unique cross layer}
\end{equation}
where $\mathcal{T}_u$ indicates the task of extracting unique representations. The configuration described in Eq. \ref{eq: unique cross layer} facilitates comprehensive interaction between modalities, enabling the attention mechanism to effectively capture unique features through cross-modal interactions. Here, each modality contributes to the attention mechanism in a way that emphasizes differences rather than commonalities, enhancing the detection of unique attributes across modal inputs.

To enhance the modality-specific fidelity of extracted unique representations, we introduce a \textit{feature reconstruction} stage, illustrated in Fig. \ref{fig: mcud}. In this stage, we employ the pretrained model $\text{H}_{fix}(\cdot)$, which has a greater number of learnable parameters compared to our backbone. The pretrained model exhibits strong representational capabilities, as it is trained on a large-scale dataset of single-modality images in a self-supervised learning manner. To align the output of our model with that of the pretrained model, we project the concatenation the unique and common representations into a latent space:
\begin{equation}
\label{multi-modal unique loss}
\begin{aligned}
    L_{m\text{-}uni}(\ebf{x}^1, \ebf{x}^2) =& \| \text{H}_{fix}(\ebf{x}^1) - \text{LP}(\ebf{f}^1_u, \ebf{f}_c) \| + \\ 
     & \| \text{H}_{fix}(\ebf{x}^2) - \text{LP}(\ebf{f}^2_u, \ebf{f}_c) \|,
\end{aligned}
\end{equation}
where $\text{LP}$ is a modality head that includes a linear layer, projecting the concatenated representations into the same latent space as that produced by the pretrained model. The feature reconstruction stage significantly refines unique representations. The pretrained model, designed for a single modality, enhances this refinement by efficiently utilizing a wide range of data within that modality. The extensive and varied nature of the large-scale data allows the pretrained model to develop more effective and efficient representations. Employing the loss function defined in (\ref{multi-modal unique loss}) during training acts as a distillation process, sharpening the unique characteristics of the representation. Further details about the application of the pretrained model across different modalities and its impact on the training process will be discussed in Sec. \ref{sec: experiments}.

\subsection{Masked Feature Model}
\label{sec: MFM}


With the assistance of the CUD tasks, the source images are decomposed into coherent common and unique representations. However, these decomposed representations are not the final output; instead, our goal is to build a network for downstream tasks that requires producing a fused image, object detection bounding boxes, or segmentation maps. Therefore, the common and unique representations serve as intermediate products. The key challenge is how to fully utilize these intermediate products to generate well-fused features from the decomposed representations, thereby better serving the downstream tasks.

To address the aforementioned challenge, we propose a masked feature model (MFM), as illustrated in Fig. \ref{fig: cud}. Inspired by the masked image model (MIM), the MFM selectively masks input tokens and requires the MFM encoder to reconstruct these tokens, aiming to predict an image that contains all necessary information. Notably, there are two significant differences between the MFM and MIM. First, the MIM implements masking at the pixel level, which inherently focuses on local spatial features. In contrast, the MFM employs masking at the feature level, enabling each token to encapsulate global information from across the entire image. Such a strategy is particularly beneficial as it avoids potential conflicts with the CUD task. Second, the MFM diverges from the completely random masking pattern typical of MIM. Considering that our representations are divided into common and unique categories, it is essential to ensure balanced exposure during training. Therefore, we implement a strategy where tokens are uniformly sampled from each category for masking. The tokens selected for masking from each category are then combined to form the input for the MFM, ensuring a comprehensive integration of the features during the reconstruction phase.


The loss function of MFM can be formulated as:
\begin{equation}
\begin{aligned}
    L&_{mfm}(\ebf{x}^1, \ebf{x}^2) = \\
    &\| \text{H}_{MFMED} \left( \texttt{Sample}\left( \ebf{f}_c, \ebf{f}_u^1, \ebf{f}_u^2 \right) \right) - \ebf{x} \|,
\end{aligned}
\end{equation}
where $\texttt{Sample}$ represents the proposed sampling strategy, designed to balance input between common and unique representations. The architecture of $\text{H}_{MFMED}$ is comprised of three key components: an encoder $\text{H}_{MFME}$, an interpolation layer, and a decoder. The interpolation layer plays a critical role by integrating masked tokens, ensuring alignment with the dimensionality requirements of the decoder. The decoder transforms the obtained representations into reconstructed images.

Combine the CUD and MFM tasks, the total loss of the self-supervised learning framework can be written as:
\begin{equation}
\hspace{-8pt} 
\begin{aligned}
 & L_{total}(\ebf{x}^1, \ebf{x}^2)= \\  & 
\begin{cases}
    \begin{aligned}
       L_{m\text{-}com}(\ebf{x}^i) + L_{m\text{-}uni}(\ebf{x}^i) & \text{ if } \ebf{x}^i \in \text{ multi-modal} &\hspace{-6pt} \text{(a)} \\
        L_{s\text{-}cud}(\ebf{x}^i) + \alpha L_{mfm}(\ebf{x}^i) & \text{ otherwise}, &\hspace{-6pt} \text{(b)}
    \end{aligned}
\end{cases}
\end{aligned}
\label{total loss}
\end{equation}
where $\alpha$ is the balance factor between the two self-supervised pretext tasks. During the training phase, both single-modal and multi-modal data are utilized. For single-modal data, the network optimization leverages the CUD and MFM tasks with loss defined in Eq. (\ref{total loss}.a). For multi-modal data, we minimize the loss of the MCUD task defined in Eq. (\ref{total loss}.b). By training with transformed views derived from both single and multi-modal data, the network acquires more general representations, enhancing its ability to adaptively handle diverse image fusion tasks.

\subsection{Downstream tasks}

\begin{figure}[htbp]
    \centering
    \includegraphics[width=9cm]{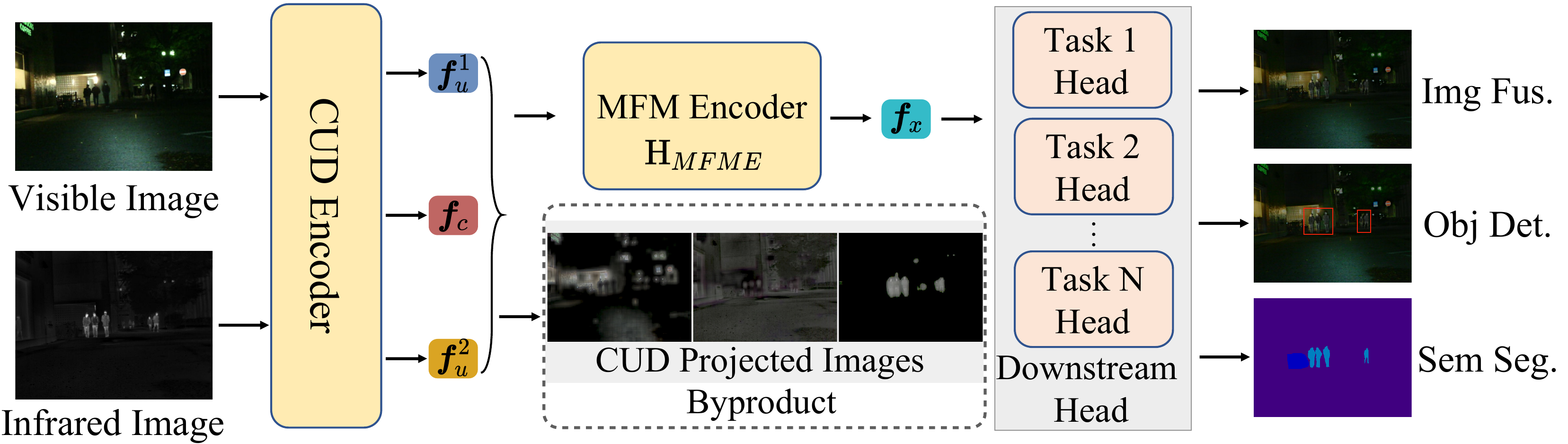}
    \caption{ The adaptation of our network framework to downstream tasks. }
    \label{fig: downstream tasks}
\end{figure}

When trained with the CUD and MFM pretext tasks, the network is designed to adeptly transition to various downstream tasks.  We focus on two representative types of tasks: low-level (image fusion) and high-level (object detection and semantic segmentation) tasks. As the network adapts to these downstream tasks, certain auxiliary outputs present during the training phase are phased out, as depicted in Fig. \ref{fig: downstream tasks}. In the following sections, we will elaborate on these tasks. For simplicity, semantic segmentation is discussed as a representative example of high-level vision tasks. The process for object detection is similar to that of semantic segmentation.

\noindent \textit{Image Fusion.} The source images $\ebf{x}^1, \ebf{x}^2$ are input into the vision transformer (ViT) encoder $\text{H}$, which extracts the features $\text{H}(\ebf{x}^1)$ and $\text{H}(\ebf{x}^2)$.  These features are then processed by the cross-attention layer, resulting in the generation of common and unique representations: $\ebf{f}_c$, $\ebf{f}_u^1$, and $\ebf{f}_u^2$. These decomposed representations are concatenated and sequentially fed into the encoder of the MFM and fused projector ($\text{PH}$) to synthesize the final fused image. For the image fusion task, the projectors of common and unique representations, the interpolation layer, the decoder of the MFM, and the pretrained large-scale encoder are not required. Instead, the fused image is directly produced using the fused projector ($\text{PH}$), which simplifies the process and allows for immediate testing without the need for further training.

\noindent \textit{Segmentation.} For the segmentation task, multi-modal images $\ebf{x}^1$ and $\ebf{x}^2$ are sequentially processed through the vision transformer (ViT) backbone $\text{H}$, the cross-attention layer $\text{CA}$, and then the MFM encoder. The MFM encoder outputs the fused features $\ebf{f}_x$. Subsequently, these features are processed by a segmentation head to produce the segmentation map. Unlike the image fusion task, the segmentation head cannot be tested immediately. Instead, the segmentation head and the backbone are required to be finetuned to optimize segmentation results. Such finetuning is a widely used strategy to verify the capability of the backbone developed through self-supervised learning.

\subsection{Visualization of (M)CUD}

\begin{figure}[htbp]
    \centering
    \begin{overpic}[width=9cm]{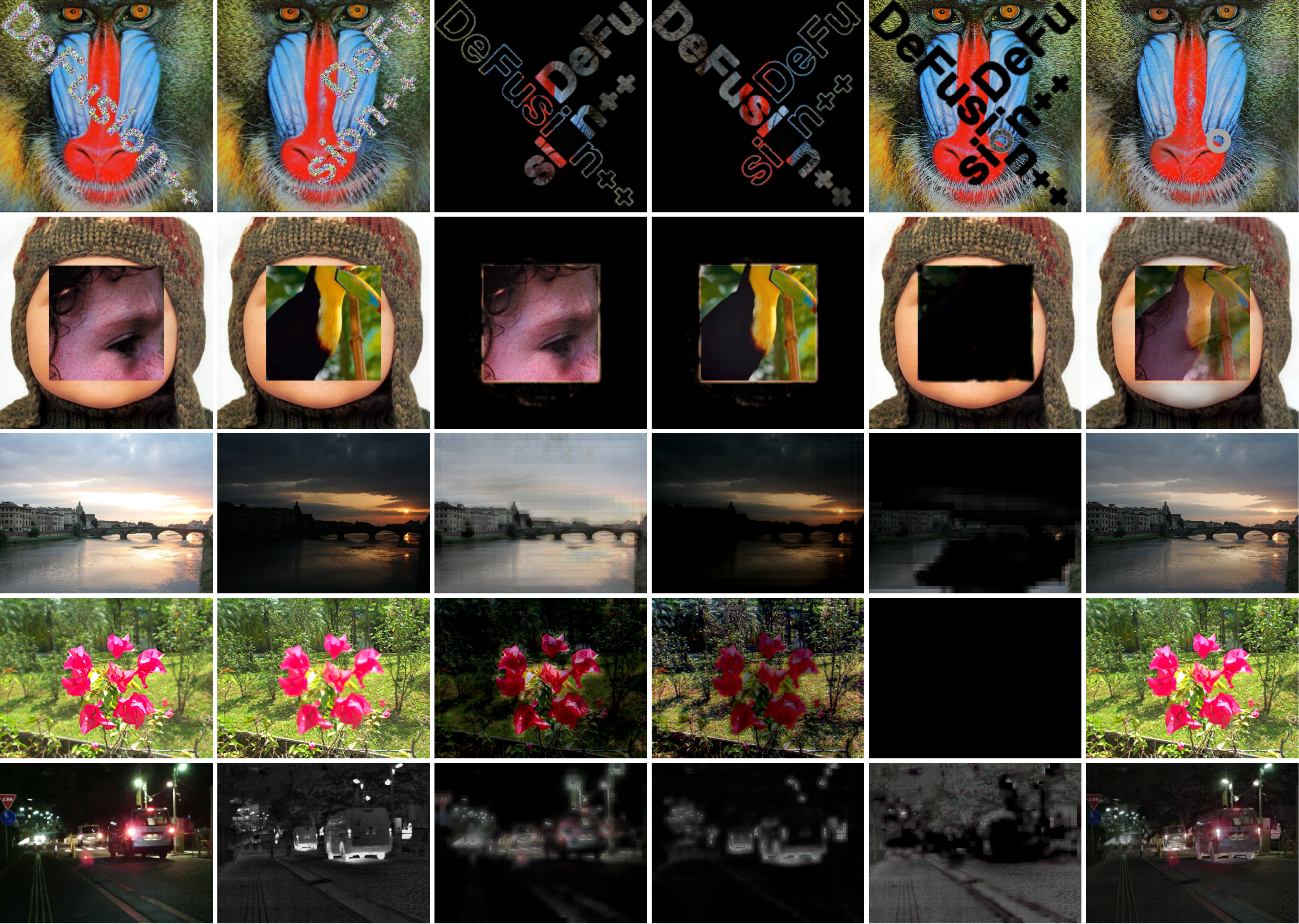}
    \put(4.9, -2.7){\color{black}\footnotesize (a)  $\ebf{x}^1$}
        \put(20.7, -2.7){\color{black}\footnotesize (b) $\ebf{x}^2$}
        \put(34.7, -2.7){\color{black}\scriptsize (c)  $\text{PH}(\ebf{f}_u^1)$}
        \put(51.0, -2.7){\color{black}\scriptsize (d)  $\text{PH}(\ebf{f}_u^2)$}
        \put(68.0, -2.7){\color{black}\scriptsize (e)  $\text{PH}(\ebf{f}_c)$}
        \put(82.0, -2.7){\color{black}\scriptsize (f) Fused Image}
    \end{overpic}
    \caption{ Visualization of common and unique decomposition results for some toy and real examples.}
    \label{fig: visualization}
\end{figure}
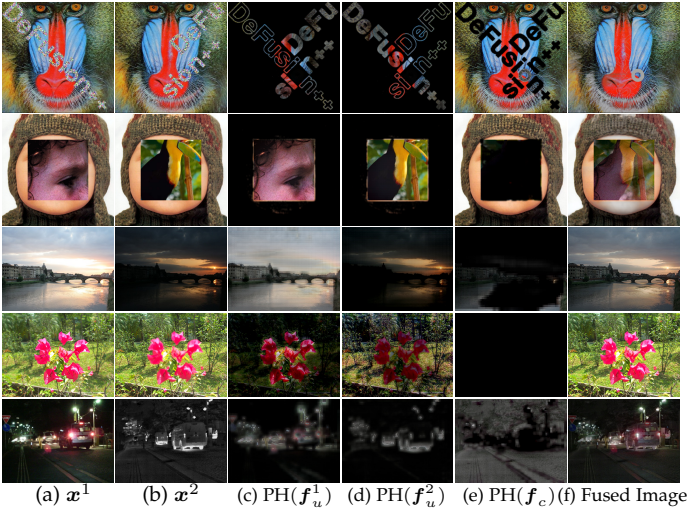

In this section, we will demonstrate the unique and common representation ability of DeFusion++. As shown in Fig. \ref{fig: visualization}, we show two toy examples in the first two rows and three image fusion task in the last three examples.

For the toy examples, we first create two specific masks that are quite different from the training setting. In the first mask, we fill it with Gaussian noise $n$. In the second toy example, we center-crop the image and replace the cropped patch with another image. After feeding the source image, DeFusion++ produces a satisfactory decomposed image; the common and unique images are accurate. Note that, in the first case, the decomposed unique image shows dilated edges, which demonstrates that DeFusion++ retains some semantic understanding capabilities. If the model had overfit to the trained task, the dilated edge characteristic should not appear. However, from the perspective of image understanding, there exists an implicit unique component, i.e., the dilated edge characteristic.

For the real examples, we selected three representative examples to demonstrate the role CUD plays in real image fusion tasks. In the MEF task, the unique component preserves the well-exposed part and discards the over-exposed regions, resulting in meaningless blank areas and dark regions from under-exposure. In the MFF task, the decomposed unique images highlight the focus region and add black masks in the defocused areas. For the common image, no focus region remains. In the IVF task, the decomposed unique visible image highlights texture details while avoiding the introduction of over-exposed content. The decomposed unique infrared image highlights regions with higher pixel intensity, which suggests the presence of an object. The decomposed common image shows the background of the scene. All these examples demonstrate that DeFusion++, trained with (M)CUD, can adaptively extract effective unique information, perfectly adapting to the image fusion task.

\section{Experiments}
\label{sec: experiments}

We evaluate DeFusion++ through a series of evaluations covering image fusion and downstream tasks. Image fusion tasks are divided into two categories: single-modal and multi-modal. Single-modal tasks include multi-exposure and multi-focus fusion, while multi-modal tasks focus on infrared and visible image fusion. For downstream tasks, we evaluate the proposed framework on semantic segmentation and object detection. To further demonstrate the effectiveness of our module and design, we also conduct ablation studies at the end of this section.

\subsection{Implementation Details} 
\noindent \textit{Network}. For the backbone of our network, we utilize the ViT-T model, which has approximately 5 M parameters. To better capture modality-specific unique features in the MCUD task, we use the ViT-B as the large feature extractor $\text{H}_{fix}$, consisting of 88 M parameters. Our MFM module is built with four block layers of the ViT-T model. Following representative SSL models \cite{he2022masked, zhou2021ibot}, the projector heads varies according to the downstream task. For image fusion, we employ a two-layer ViT projector. For segmentation tasks, a conventional upsampling head is used, while object detection tasks utilize an RCNN Head \cite{girshick2014rich}.

\noindent \textit{Dataset}. For the CUD task, we apply the degradation transformations (Eq. \ref{decomposition equation}) to the COCO dataset to create two views of each image. For the MUCD task, we train the model using the MSRS \cite{tang2022piafusion} dataset, which is specifically tailored for the IVF task. To further improve the robustness of the pretrained feature extractor $\text{H}_{fix}$ to infrared modality, we also pretrain it on the domain-specific FLIR\footnote{https://www.flir.eu/oem/adas/adas-dataset-form/} dataset.

\noindent \textit{Training Details}. We crop the source images into patches of 224 $\times$ 224 pixels. We apply random flipping and color jitter to the source images to enhance data variability. We train our model for 30 epochs with an initial learning rate of 1e-4, which decays to half after 20 epochs. The model is optimized using the Adam optimizer with a batch size of 16. For the hyperparameters, we set $\alpha$ in Eq. \ref{total loss} to 0.1.

\subsection{Single Modality Image Fusion}
\subsubsection{Multi-exposure Fusion}

\begin{figure*}[htbp]
    \centering
    \begin{overpic}[width=18cm]{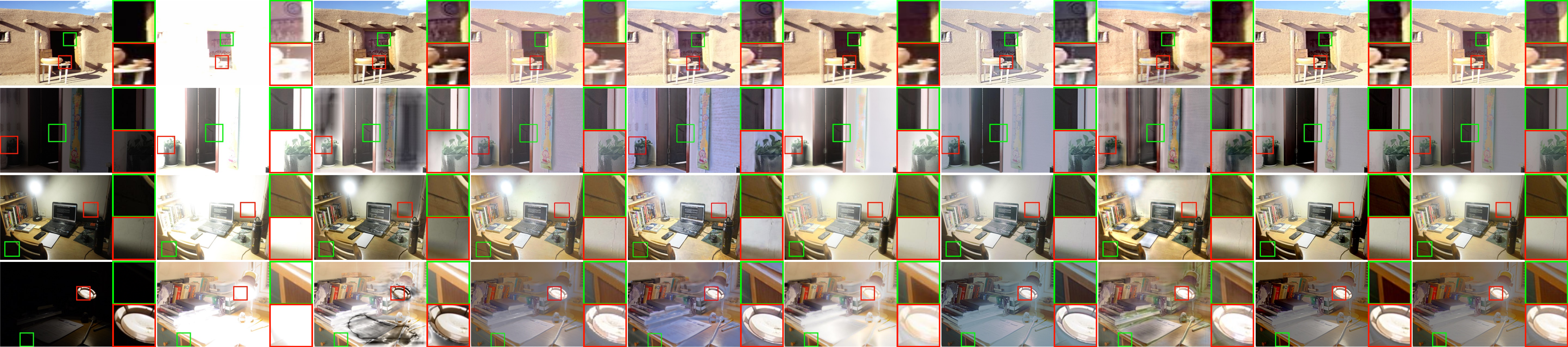}
    \put(1.3, -1.5){\color{black}\footnotesize (a) Under}
        \put(11.7, -1.5){\color{black}\footnotesize (b) Over}
        \put(20.8, -1.5){\color{black}\footnotesize (c) MEFNet}
        \put(30.7, -1.5){\color{black}\footnotesize (d) HoLoCo}
        \put(41.1, -1.5){\color{black}\footnotesize (e) CRMEF}
        \put(50.6, -1.5){\color{black}\footnotesize (f) MEFLUT}
        \put(61.2, -1.5){\color{black}\footnotesize (g) IFCNN}
        \put(71.3, -1.5){\color{black}\footnotesize (h) CUNet}
        \put(80.3, -1.5){\color{black}\footnotesize (i) U2Fusion}
        \put(89.8, -1.5){\color{black}\footnotesize (j) DeFusion++}
    \end{overpic}
    \caption{Qualitative comparison of our DeFusion++ with seven MEF methods on four over- and under- exposure image pairs on the MEFB dataset. }
    \label{fig: mef-mefb}
\end{figure*}

\begin{figure*}[htbp]
    \centering
    \begin{overpic}[width=18cm]{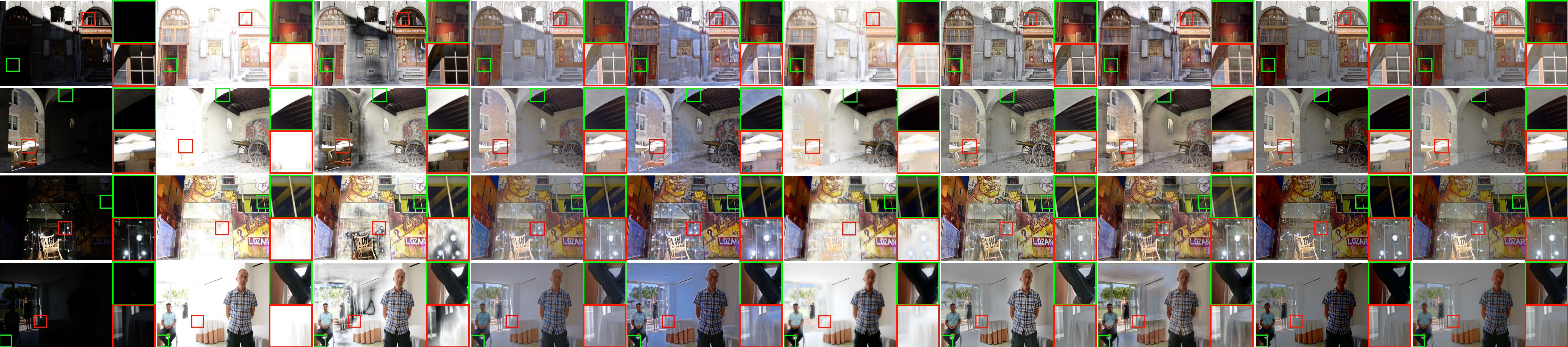}
    \put(1.3, -1.5){\color{black}\footnotesize (a) Under}
        \put(11.7, -1.5){\color{black}\footnotesize (b) Over}
        \put(20.8, -1.5){\color{black}\footnotesize (c) MEFNet}
        \put(30.7, -1.5){\color{black}\footnotesize (d) HoLoCo}
        \put(41.1, -1.5){\color{black}\footnotesize (e) CRMEF}
        \put(50.6, -1.5){\color{black}\footnotesize (f) MEFLUT}
        \put(61.2, -1.5){\color{black}\footnotesize (g) IFCNN}
        \put(71.3, -1.5){\color{black}\footnotesize (h) CUNet}
        \put(80.3, -1.5){\color{black}\footnotesize (i) U2Fusion}
        \put(89.8, -1.5){\color{black}\footnotesize (j) DeFusion++}
    \end{overpic}
    \caption{Qualitative comparison of our DeFusion++ with seven MEF methods on four over- and under- exposure image pairs on the SICE dataset. }
    \label{fig: mef-sice}
\end{figure*}

In our comparison, DeFusion++ is benchmarked against nine state-of-the-art (SOTA) image fusion methods: MEFNet \cite{ma2019mefnet}, HoLoCo \cite{liu2023holoco}, CRMEF \cite{liu2023embracing}, MEFLUT \cite{jiang2023meflut}, HSDS \cite{wu2024hybrid}, IFCNN \cite{zhang2020ifcnn}, CUNet \cite{deng2020deep}, PMGI \cite{zhang2020rethinking}, and U2Fusion \cite{xu2020u2fusion}. Among these methods, MEFNet, HoLoCo, CRMEF, MEFLUT and HSDS are tailored for the MEF task, whereas the remaining are applicable to a variety of fusion tasks. For assessment, we employ the MEFB \cite{zhang2021benchmarking} and SICE \cite{cai2018learning} datasets. MEFB, a comprehensive hybrid dataset, comprises 100 image pairs drawn from multiple sources, including the EMPA\footnote{http://www.empamedia.ethz.ch/hdrdatabase/index.php} database along with other distinct datasets \cite{zeng2014perceptual, ram2017deepfuse}. The SICE dataset, considerably larger, contains 589 image pairs, providing a variety of lighting and exposure conditions for testing. The qualitative results are illustrated in Figs. \ref{fig: mef-mefb} and \ref{fig: mef-sice}, with the quantitative performance comparisons detailed in Tab. \ref{tab: mef_performance}.

For the quantitative evaluation, we select four metrics to assess the quality of image fusion: nonlinear correlation information entropy (NCIE) \cite{wang2008performance}, which quantifies correlation coefficient in fused images; edge-based similarity measurement ($N^{AB/F}$), reflecting the presence of noise or artifacts; structural similarity index measure (SSIM); and correlation coefficient (CC). Additionally, the multi-exposure image structural similarity index measure (MEF-SSIM) \cite{ma2015perceptual}, specifically designed for multi-exposure fusion, is employed. DeFusion++ consistently demonstrates high performance across these metrics, as detailed in Tab. \ref{tab: mef_performance}. Notably, DeFusion++ minimizes noise and artifacts, achieving the highest scores in $N^{AB/F}$. Our method achieves the highest performance on both the MEFB and SICE datasets, as indicated by the MEF-SSIM, a metric tailored for the MEF task. These results demonstrate the superior capability of DeFusion++ over other SOTA methods in two significant benchmarks, thus confirming its efficacy in MEF tasks.

\begin{table*}
\centering
\caption{Performance comparison of MEF across datasets MEFB and SICE. Best values are in \textbf{bold} and second-best values are \underline{underlined}.}
\label{tab: mef_performance}
\begin{tabular}{l ccccccccccc}
\toprule
\multicolumn{2}{c}{\multirow{2}{*}{Method}} & \multicolumn{5}{c}{MEFB} & \multicolumn{5}{c}{SICE} \\ \cmidrule(lr){3-7} \cmidrule(lr){8-12}
                     &   & NCIE$\uparrow$ & $N^{AB/F}\downarrow$ & SSIM$\uparrow$ & CC$\uparrow$ & MEF-SSIM$\uparrow$ & NCIE$\uparrow$ & $N^{AB/F}\downarrow$ & SSIM$\uparrow$ & CC$\uparrow$ & MEF-SSIM$\uparrow$ \\ \midrule
                        
\color{dt} MEFNet  \cite{ma2019mefnet} & \color{dt} TIP'19                & \color{dt} 0.8152 & \color{dt} 0.0285 & \color{dt} \textbf{1.1845} & \color{dt} 0.5254 & \color{dt} 0.7958 & \color{dt} 0.8108 & \color{dt} 0.0593 & \color{dt} 0.9903 & \color{dt} 0.3193 & \color{dt} 0.6825 \\
\color{dt} HoLoCo \cite{liu2023holoco} & \color{dt} IF'23 & \color{dt} 0.8116 & \color{dt} 0.0430 & \color{dt} 1.0417 & \color{dt} 0.8678 & \color{dt} 0.7451 & \color{dt} 0.8081 & \color{dt} 0.0380 & \color{dt} 0.9032 & \color{dt} 0.8195 & \color{dt} 0.7310 \\
\color{dt} CRMEF \cite{liu2023embracing} & \color{dt} TCSVT'24 & \color{dt} 0.8118 & \color{dt} 0.0536 & \color{dt} 1.1382 & \color{dt} 0.8694 & \color{dt} 0.8065 & \color{dt} 0.8075 & \color{dt} 0.0546 & \color{dt} 0.9712 & \color{dt} 0.7990 & \color{dt} 0.7580 \\
\color{dt} MEFLUT  \cite{jiang2023meflut} & \color{dt} ICCV'23 & \color{dt} 0.8128 & \color{dt} \underline{0.0098} & \color{dt} 1.0778 & \color{dt} 0.6287 & \color{dt} 0.7824 & \color{dt} 0.8110 & \color{dt} \underline{0.0074} & \color{dt} \textbf{1.1017} & \color{dt} 0.7346 & \color{dt} 0.7821 \\
\color{dt} HSDS \cite{wu2024hybrid} & \color{dt} AAAI'24 & \color{dt} 0.8139 & \color{dt} 0.0296 & \color{dt} 1.1314 & \color{dt} 0.8843 & \color{dt} 0.8046 & \color{dt} 0.8092 & \color{dt} 0.0508 & \color{dt} 0.9598 & \color{dt} 0.8293 & \color{dt} 0.7558 \\
CUNet   \cite{deng2020deep}    & PAMI'20  & 0.8096 & 0.0404 & 1.0943 & 0.8292 & 0.7937 & 0.8064 & 0.0577 & 0.9299 & 0.7538 & 0.7391 \\
IFCNN   \cite{zhang2020ifcnn}  & IF'20 & 0.8151 & 0.0495  & 1.1456 & 0.8921 & 0.8175 & 0.8060 & 0.1315 & 0.9421 & 0.7255 & 0.6953 \\
PMGI  \cite{zhang2020rethinking}  & AAAI'20 & \underline{0.8171} & 0.0475 & 1.0944 & \underline{0.9017} & \underline{0.8217} & 0.8109 & 0.0779 & 0.9068 & 0.8204 & 0.7356 \\
U2Fusion  \cite{xu2020u2fusion}   & PAMI'22 & 0.8161 & 0.0691 & 1.0710 & 0.8973 & 0.8151 & \underline{0.8113} & 0.0245 & 0.9385 & \underline{0.8424} & \underline{0.7959} \\
DeFusion++        &      & \textbf{0.8200} & \textbf{0.0065} & \underline{1.1619} & \textbf{0.9068} & \textbf{0.8407} & \textbf{0.8137} & \textbf{0.0057} & \underline{1.0249} & \textbf{0.8602} & \textbf{0.8054} \\
\bottomrule
\end{tabular}
\begin{tablenotes}
\item We use \textcolor{dt}{gray} to mark methods specifically designed for the MEF task.
\end{tablenotes}
\end{table*}

As shown in Fig. \ref{fig: mef-mefb}, two representative types of patches are extracted and zoom-in from each example to precisely evaluate the effectiveness of different methods in handling challenging scenes. The first type of patch primarily shows detailed information from the under-exposed image, while the corresponding patch from the over-exposed image contains minimal information. Conversely, the second type is rich in details from the over-exposed image, while the corresponding patch from the under-exposed image provides limited information. Several methods, including MEFNet, CRMEF, MEFLUT, IFCNN, and CUNet, exhibit significant limitations, particularly evident in the fourth example. The results show a lack of balance in luminance and color consistency. Specifically, CRMEF and IFCNN produce an unnatural pseudo-blue/purple color shift, while MEFNet, MEFLUT, and CUNet introduce exposure-related artifacts across the image. Focusing on the cropped patches, it becomes evident that methods such as HoLoCo fail to preserve valuable information from the over-exposed images and introduce spatial aliasing artifacts. Similarly, U2Fusion exhibits difficulties in capturing rich textures from under-exposed images, as highlighted in the second example. In contrast, DeFusion++ consistently achieves an optimal balance, retaining faithful details between the over-exposed and under-exposed images.

The fused results on the SICE dataset, as shown in Fig. \ref{fig: mef-sice}, which comprises higher resolution images than those in the MEFB, present a greater challenge for fusion methods in effectively handling these scenes. Similar to the results from the MEFB, MEFNet, CRMEF, and MEFLUT produce artifacts and color inconsistencies. In the cropped patches shown in Fig. \ref{fig: mef-sice}, it is evident that methods such as MEFLUT and U2Fusion struggle to retain details from over-exposed images, as highlighted by the green box in the second example. Furthermore, CUNet fails to preserve sharp edges in the fused images, as illustrated in the red box in the same example. In contrast, DeFusion++ produces fused images that appear natural and avoid introducing color artifacts. In the zoomed-in regions, the images generated by DeFusion++ preserve the richest details from both over- and under-exposed images, demonstrating its superiority and adaptability across various scenes.

\subsubsection{Multi-focus Fusion}

\begin{figure*}[htbp]
    \centering
    \begin{overpic}[width=18cm]{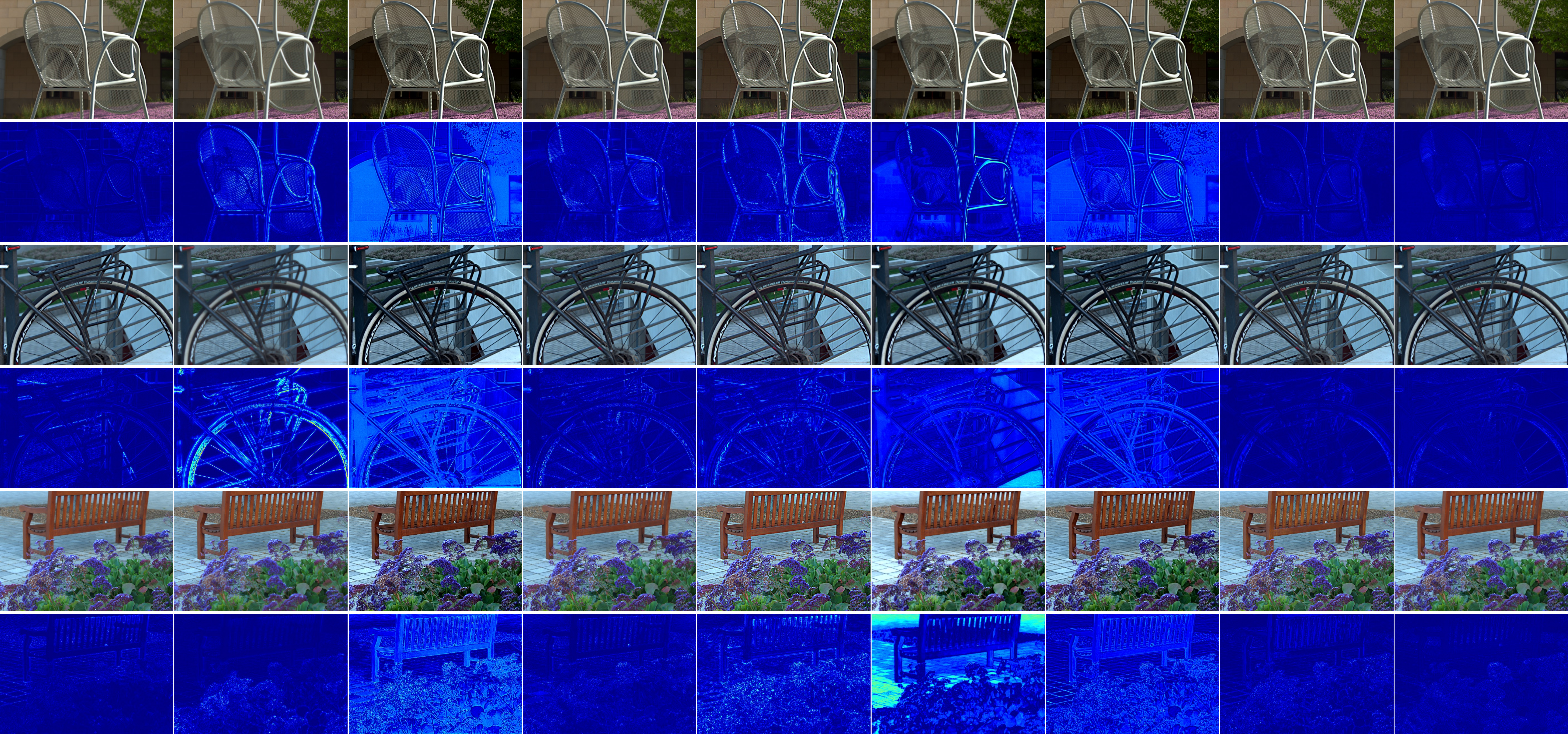}
    \put(3.0, -1.5){\color{black}\footnotesize (a) Near}
        \put(13.7, -1.5){\color{black}\footnotesize (b) Far}
        \put(22.8, -1.5){\color{black}\footnotesize (c) MFFGAN}
        \put(35.0, -1.5){\color{black}\footnotesize (d) ZMFF}
        \put(46.0, -1.5){\color{black}\footnotesize (e) CUNet}
        \put(57.7, -1.5){\color{black}\footnotesize (f) PMGI}
        \put(67.8, -1.5){\color{black}\footnotesize (g) U2Fusion}
        \put(79.5, -1.5){\color{black}\footnotesize (h) IFCNN}
        \put(89.2, -1.5){\color{black}\footnotesize (i) DeFusion++}
    \end{overpic}
    \caption{Qualitative comparison of DeFusion++ with six MFF methods on six multi-focus image pairs on the RealMFF dataset. We provide the residual maps for each result of comparison and input images to highlight the difference with GT.}
    \label{fig: mff-mfif}
\end{figure*}
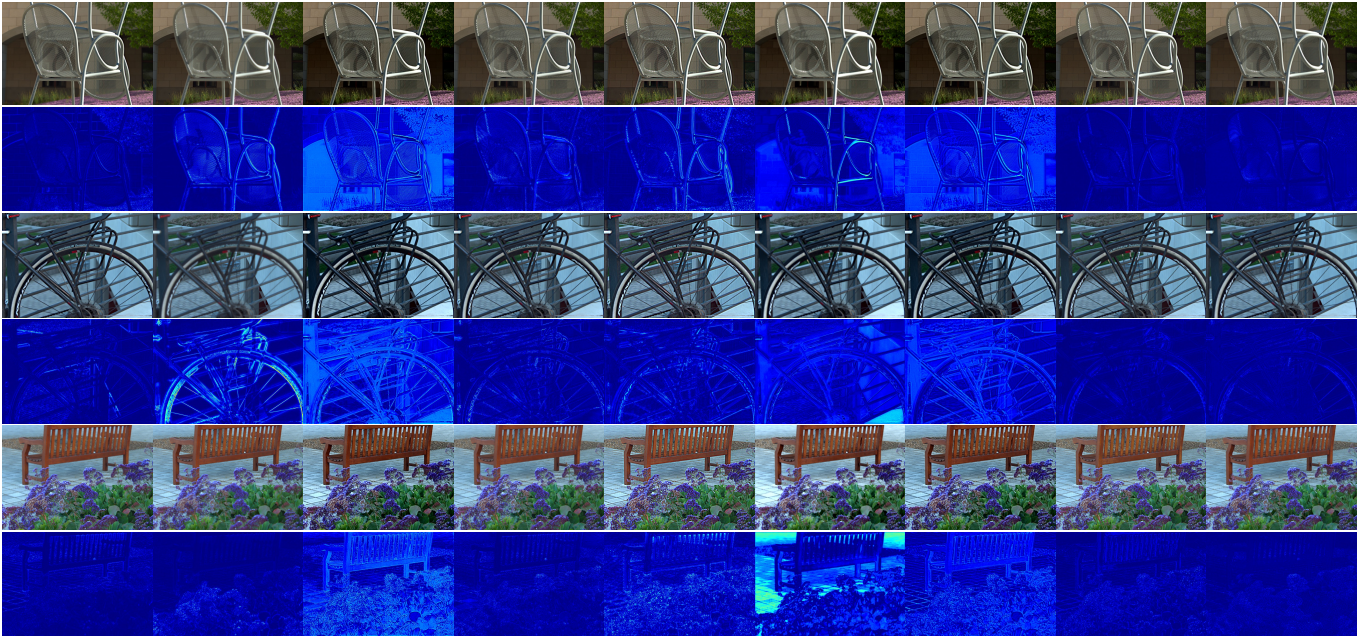

We compare DeFusion++ with six SOTA methods: MFFGAN \cite{zhang2021mff}, ZMFF \cite{hu2023zmff}, CUNet \cite{deng2020deep}, IFCNN \cite{zhang2020ifcnn}, PMGI \cite{zhang2020rethinking}, and U2Fusion \cite{xu2020u2fusion}. MFFGAN and ZMFF are specifically designed for the MFF task, whereas the other methods are general image fusion approaches. We evaluate these MFF methods on the MFIF dataset, which comprises images captured with a light-field camera, including an all-in-focus reference image for each scene. The qualitative results are presented in Fig. \ref{fig: mff-mfif}. The quantitative comparison performances, including those on the MFIF dataset \cite{zhang2021mfif} and the RealMFF dataset \cite{zhang2020real}, are shown in Tab. \ref{tab: mff_performance_comparison}. Notably, the MFIF dataset provides only the degraded blurred images without the corresponding ground truth.

\begin{table}[htbp]
    \centering
    \setlength{\tabcolsep}{2pt} 
    \begin{threeparttable}
        \caption{Performance comparison of MFF across datasets MFIF and RealMFF.}
        \label{tab: mff_performance_comparison}
        \begin{tabular}{l@{\hspace{0pt}}c@{\hspace{4pt}}cc@{\hspace{4pt}}cc@{\hspace{4pt}}cc}
        \toprule
        \multicolumn{2}{c}{\multirow{2}{*}{Method}} & \multicolumn{2}{c}{MFIF\tnote{1}} & \multicolumn{2}{c}{RealMFF\tnote{1}} & \multicolumn{2}{c}{RealMFF} \\
        \cmidrule(lr){3-4} \cmidrule(lr){5-6} \cmidrule(lr){7-8}
        &  & PSNR$\uparrow$ & SSIM$\uparrow$ & PSNR$\uparrow$ & SSIM$\uparrow$ & PSNR$\uparrow$ & SSIM$\uparrow$ \\
        \midrule
        \color{dt} MFFGAN \cite{zhang2021mff} & \color{dt}IF'21 & \color{dt} 24.3020 & \color{dt} 0.8792 & \color{dt} 22.8049 & \color{dt} 0.8106 & \color{dt} 24.1350 & \color{dt} 0.8496 \\
        \color{dt} ZMFF \cite{hu2023zmff} & \color{dt}IF'23 & \color{dt} 19.1561 & \color{dt} 0.7929 & \color{dt} 31.2874  & \color{dt} 0.9575 & \color{dt} 31.9893  & \color{dt} \underline{0.9644} \\
        CUNet \cite{deng2020deep} & PAMI'20 & 24.8838 & 0.8740 & 26.6580 & 0.8994 & 29.1651 & 0.9381 \\
        IFCNN \cite{zhang2020ifcnn} & IF'20 & \underline{26.9077} & \underline{0.9046} & \underline{32.9256} & \underline{0.9641} & \textbf{36.9205} & \textbf{0.9829} \\
        PMGI \cite{zhang2020rethinking} & AAAI'20 & 20.8805 & 0.8651 & 24.0881 & 0.8898 & 24.6635 & 0.9032 \\
        U2Fusion \cite{xu2020u2fusion} & PAMI'22 & 21.9414 & 0.8148 & 23.9866 & 0.8487 & 25.2597 & 0.8795 \\
        DeFusion++ & & \textbf{27.8805} & \textbf{0.9188} & \textbf{33.4091} & \textbf{0.9659} & \underline{32.0084} & {0.9534} \\
        \bottomrule
        \end{tabular}
        \begin{tablenotes}
        \item[1] No-reference conditions.
        \end{tablenotes}
    \end{threeparttable}
\end{table}

The quantitative evaluation results are present in Tab. \ref{tab: mff_performance_comparison}. MFF holds unique importance in image fusion tasks because it theoretically allows for the existence of a ground truth, \ie all-in-focus image. Accordingly, we employ two evaluation criteria to compare methods: firstly, by assessing the quality of the fused image against the source images, and secondly, by directly scoring the fused image against the ground truth. Under the first criterion, DeFusion++ demonstrates the best performance across different datasets. Under the second criterion, our method achieves the second-best performance in terms of PSNR. It is acceptable, considering that IFCNN, which uses ground truth during training as a supervised learning method, secures the top position.

To better visualize the disparity in the fused images of MFF tasks, we present the residual map for each image, as illustrated in Fig. \ref{fig: mff-mfif}. The residual map reveals the disparity between the degraded/recovered image and the ground truth. Significantly, our method and IFCNN exhibit a clear advantage compared to their counterparts. It is noteworthy that IFCNN is a supervised model, developed through training with pairs of blurred and sharp images. In contrast, our method achieves strong performance without incorporating specific considerations tailored for the MFF task.

\subsection{Multiple Modality Image Fusion}

\begin{figure*}[htbp]
    \centering
    \begin{overpic}[width=18cm]{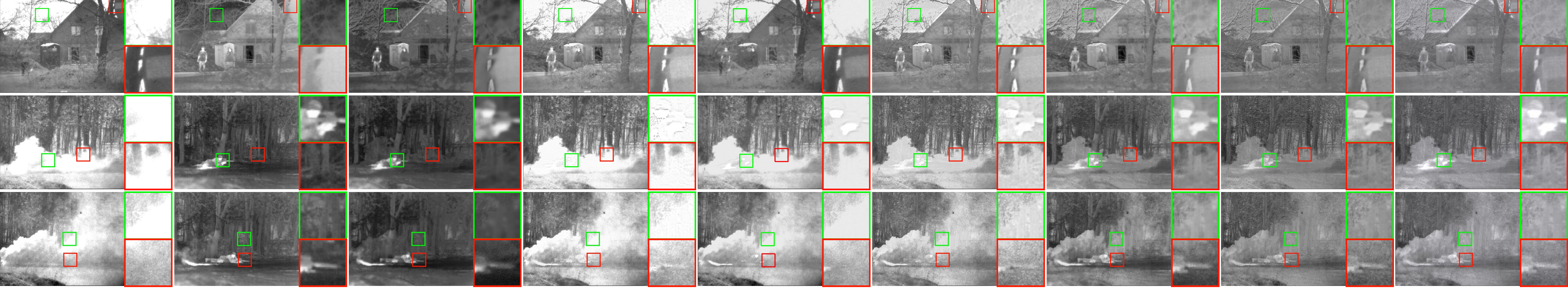}
        \put(1.7, -1.5){\color{black}\footnotesize (a) Visible}
        \put(12.4, -1.5){\color{black}\footnotesize (b) Infrared}
        \put(22.3, -1.5){\color{black}\footnotesize (c) FusionGAN}
        \put(33.9, -1.5){\color{black}\footnotesize (d) CDDFuse}
        \put(45.0, -1.5){\color{black}\footnotesize (e) TIMFusion}
        \put(57.1, -1.5){\color{black}\footnotesize (f) IFCNN}
        \put(68.7, -1.5){\color{black}\footnotesize (g) PMGI}
        \put(78.5, -1.5){\color{black}\footnotesize (h) U2Fusion}
        \put(89.0, -1.5){\color{black}\footnotesize (i) DeFusion++}

    \end{overpic}

    \caption{Qualitative comparison of DeFusion++ with six IVF methods on three representative visible and infrared image pairs on the TNO dataset. }
    \label{fig: ivf-tno}
\end{figure*}

\begin{figure*}[htbp]
    \centering
    \begin{overpic}[width=18cm]{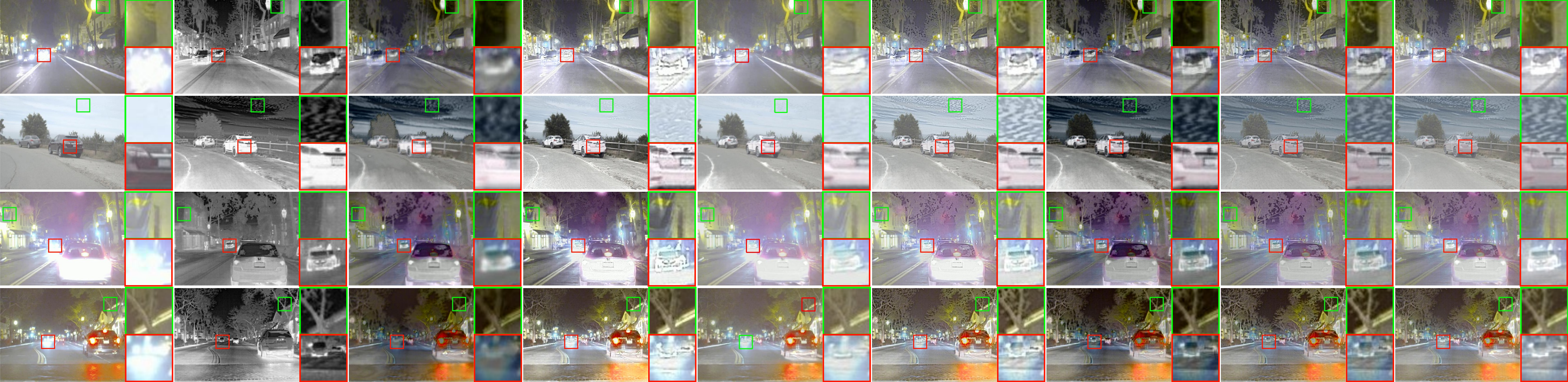}
        \put(1.7, -1.5){\color{black}\footnotesize (a) Visible}
        \put(12.4, -1.5){\color{black}\footnotesize (b) Infrared}
        \put(22.3, -1.5){\color{black}\footnotesize (c) FusionGAN}
        \put(33.9, -1.5){\color{black}\footnotesize (d) CDDFuse}
        \put(45.0, -1.5){\color{black}\footnotesize (e) TIMFusion}
        \put(57.1, -1.5){\color{black}\footnotesize (f) IFCNN}
        \put(68.7, -1.5){\color{black}\footnotesize (g) PMGI}
        \put(78.5, -1.5){\color{black}\footnotesize (h) U2Fusion}
        \put(89.0, -1.5){\color{black}\footnotesize (i) DeFusion++}
    \end{overpic}
    \caption{Visual comparison of DeFusion++ against six SOTA IVF methods on four visible and infrared image pairs from the RoadScene dataset.}
    \label{fig: ivf-roadscene}
\end{figure*}

\begin{figure*}[htbp]
    \centering
    \begin{overpic}[width=18cm]{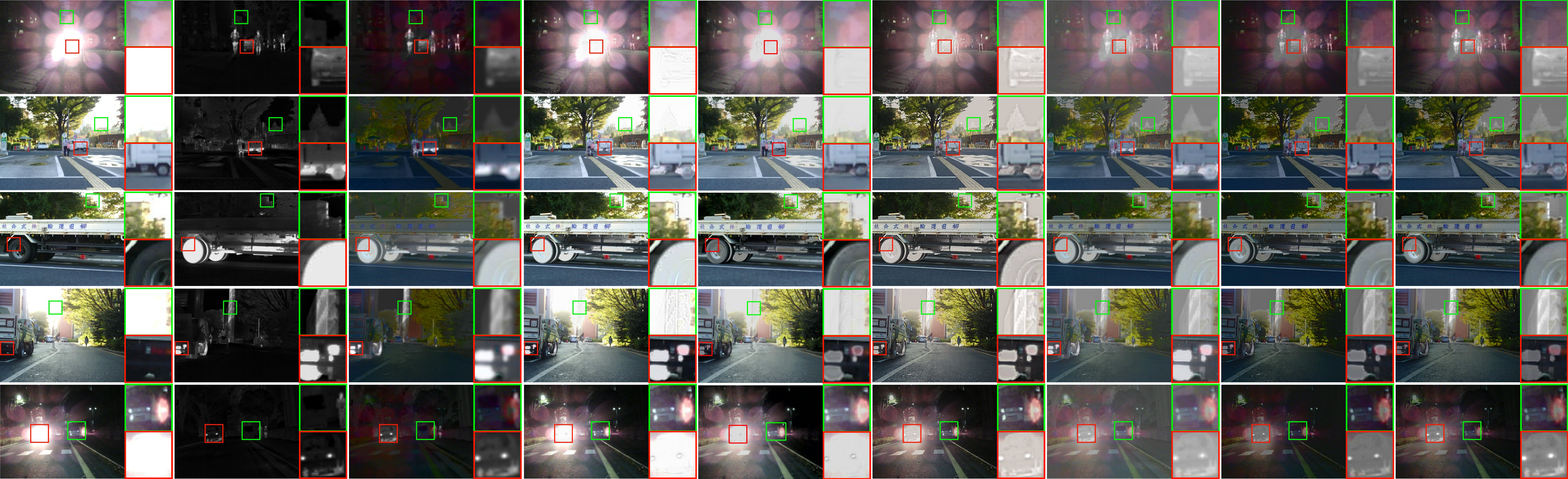}
        \put(1.7, -1.5){\color{black}\footnotesize (a) Visible}
        \put(12.4, -1.5){\color{black}\footnotesize (b) Infrared}
        \put(22.3, -1.5){\color{black}\footnotesize (c) FusionGAN}
        \put(33.9, -1.5){\color{black}\footnotesize (d) CDDFuse}
        \put(45.0, -1.5){\color{black}\footnotesize (e) TIMFusion}
        \put(57.1, -1.5){\color{black}\footnotesize (f) IFCNN}
        \put(68.7, -1.5){\color{black}\footnotesize (g) PMGI}
        \put(78.5, -1.5){\color{black}\footnotesize (h) U2Fusion}
        \put(89.0, -1.5){\color{black}\footnotesize (i) DeFusion++}
    \end{overpic}
    \caption{Qualitative analysis of DeFusion++ against six IVF methods using five visible and infrared image pairs from the MSRS dataset.
    }
    \label{fig: ivf-msrs}
\end{figure*}

To evaluate the effectiveness of the proposed method for multiple modality image fusion, we compare DeFusion++ with seven SOTA infrared visible fusion methods: FusionGAN \cite{ma2019fusiongan}, CDDFuse \cite{zhao2023cddfuse}, MetaFusion \cite{zhao2023metafusion}, TIMFusion \cite{liu2024task}, IFCNN \cite{zhang2020ifcnn}, PMGI \cite{zhang2020rethinking}, and U2Fusion \cite{xu2020u2fusion}. Among these, FusionGAN, CDDFuse, MetaFusion, and TIMFusion are specifically designed for the infrared and visible fusion (IVF) task, while the remaining methods are general image fusion approaches. To comprehensively evaluate the performance of these methods, we employ three datasets, namely TNO \cite{Toet2014}, RoadScene \cite{xu2020u2fusion}, and MSRS \cite{tang2022piafusion}, to test the robustness of each method across various scenes. The visual results are depicted in Figs. \ref{fig: ivf-tno}, \ref{fig: ivf-roadscene}, and \ref{fig: ivf-msrs}. The quantitative outcomes are presented in Tab. \ref{tab: ivf comparison}.

Quantitative comparisons are presented in Tab. \ref{tab: ivf comparison}. To evaluate the performance of the methods, four metrics are utilized: $N^{AB/F}$, SSIM, CC, and a deep-learning-based metric, CLIPIQA \cite{wang2023exploring}. CLIPIQA assesses both the quality of images (sharpness, brightness, and contrast) and abstract qualities (naturalness, scariness). As demonstrated in Tab. \ref{tab: ivf comparison}, DeFusion++ surpasses the other methods across all metrics and datasets. Notably, our method shows a significant improvement on the CLIPIQA metric, with a gain of 0.0186 on the MSRS dataset. The MSRS dataset features the most diverse scenes under various illumination conditions, highlighting the robustness and superior fusion performance of the proposed method.

\begin{table*}[htbp]
\centering
\caption{Performance comparison of IVF across datasets MSRS, TNO, and RoadScene.}
\label{tab: ivf comparison}
\setlength{\tabcolsep}{2.6pt} 
\begin{tabular}{lccccccccccccc}
\toprule
\multicolumn{2}{c}{\multirow{2}{*}{Method}} & \multicolumn{4}{c}{\textbf{MSRS}} & \multicolumn{4}{c}{\textbf{TNO}} & \multicolumn{4}{c}{\textbf{RoadScene}} \\ \cmidrule(lr){3-6} \cmidrule(lr){7-10} \cmidrule(lr){11-14}
\multicolumn{1}{c}{} & & $N^{AB/F}\downarrow$ & SSIM$\uparrow$ & CC$\uparrow$ & CLIPIQA$\uparrow$ & $N^{AB/F}\downarrow$  & SSIM$\uparrow$ & CC$\uparrow$ & CLIPIQA$\uparrow$ & $N^{AB/F}\downarrow$  & SSIM$\uparrow$ & CC$\uparrow$ & CLIPIQA$\uparrow$ \\ \midrule
\color{dt} FusionGAN \cite{ma2019fusiongan} & \color{dt} IF'19 & \color{dt} 0.0084 & \color{dt} 1.1774 & \color{dt} 0.5981 & \color{dt} 0.0997 & \color{dt} 0.0181 & \color{dt} 1.2448 & \color{dt} 0.4777 & \color{dt} 0.1863 & \color{dt} 0.0165 & \color{dt} 1.1903 & \color{dt} 0.5612 & \color{dt} 0.1177 \\
\color{dt} CDDFuse \cite{zhao2023cddfuse} & \color{dt} CVPR'23 & \color{dt} 0.0241 & \color{dt} 1.3671 & \color{dt} 0.6045 & \color{dt} 0.1557 & \color{dt} 0.0479 & \color{dt} 1.3726 & \color{dt} 0.4786 & \color{dt} \underline{0.2227} & \color{dt} 0.0883 & \color{dt} 1.3344 & \color{dt} 0.6163 & \color{dt} 0.1474\\
\color{dt} MetaFusion \cite{zhao2023metafusion} & \color{dt} CVPR'23 &  \color{dt} 0.0815 & \color{dt} 1.2262 & \color{dt} 0.6047 & \color{dt} \underline{0.1876} & \color{dt} 0.2127 & \color{dt} 1.1466 & \color{dt} 0.5172 & \color{dt} \underline{0.2227} & \color{dt} 0.1528 & \color{dt} 1.2055 & \color{dt} 0.6043 & \color{dt} 0.1538\\
\color{dt} TIMFusion \cite{liu2024task} & \color{dt} PAMI'24 & \color{dt} 0.0110 & \color{dt} 0.9982 & \color{dt} 0.5261 & \color{dt} 0.1267 & \color{dt} \underline{0.0071} & \color{dt} 1.3671 & \color{dt} 0.4059 & \color{dt} {0.2050} & \color{dt} \underline{0.0106} & \color{dt} 1.3648 & \color{dt} 0.5593 & \color{dt} \underline{0.1666} \\
IFCNN \cite{zhang2020ifcnn} & IF'20 & 0.0197 & \underline{1.4263} & 0.6327 & {0.1619} & 0.0377 & 1.3966 & 0.5447 & {0.2051} & 0.0283 & 1.4136 & 0.6269 & {0.1632} \\
PMGI \cite{zhang2020rethinking} & AAAI'20 & 0.0155 & 0.8778 & \underline{0.6496} & 0.1398 & 0.0297 & 1.3753 & 0.5518 & 0.1854 & 0.0169 & 1.3350 & 0.5906 & 0.1308 \\
U2Fusion \cite{xu2020u2fusion} & PAMI'22 & \underline{0.0054} & 1.3229 & 0.6231 & 0.1591 & {0.0166} & \underline{1.4429} & \underline{0.5777} & {0.2198} & {0.0132} & \underline{1.4456} & \underline{0.6348} & 0.1564 \\
DeFusion++ & & \textbf{0.0008}  & \textbf{1.4462} & \textbf{0.6564} & \textbf{0.2062} & \textbf{0.0032} & \textbf{1.4522} & \textbf{0.5900} & \textbf{0.2281} & \textbf{0.0036} & \textbf{1.4792} & \textbf{0.6661} & \textbf{0.2427}\\ \bottomrule
\end{tabular}%
\end{table*}

As depicted in Fig. \ref{fig: ivf-tno}, comparison methods, notably CDDFuse and TIMFusion, struggle to distinguish useful information, resulting in fog retention within the fused images, as evidenced in the second and third examples. Similarly, FusionGAN exhibits a tendency toward blurriness in fused images, a degradation observable in Fig. \ref{fig: ivf-roadscene}. Methods such as IFCNN, PMGI, and U2Fusion manage to retain essential information in fused images; however, objects within these images lack prominent salience, as highlighted by the red box in the first example of Fig. \ref{fig: ivf-msrs}. In contrast, DeFusion++ excels at preserving useful information accurately. For instance, in the last example of Fig. \ref{fig: ivf-msrs}, the intensity from the infrared spectrum is carefully preserved (as indicated by the red box), while the intensity from the visible spectrum, which may lead to over-exposure, is suppressed. Such adept adaptation highlights the superior performance of DeFusion++ in recognizing the importance of infrared intensity for object detection, such as identifying a car. It also effectively prevents important details from being obscured by over-exposure from the visible spectrum. Comprehensive qualitative comparisons demonstrate the exceptional fusion performance of DeFusion++, especially in challenging conditions marked by nighttime, fog, occlusion, and overexposure.

\subsection{Downstream Tasks}

\subsubsection{Object Detection}

We conduct object detection experiments on the M3FD dataset \cite{liu2022target}, which comprises 4,200 pairs of infrared and visible images across six classes. The dataset is divided into training, validation, and test sets in an 8:1:1 ratio, adhering to the settings specified in \cite{zhao2023cddfuse}. To ensure a competitive evaluation, we use the ViTDeT object detection framework \cite{li2022exploring}, which has a vision transformer base (ViT-B) backbone, to train the comparison methods. Our method, however, uses a ViT-T backbone with fewer learnable parameters. The setting gives the comparison methods an advantage, as ViT-B is the smallest available model in ViTDeT. All methods, including ours, employ the same ROIHead \cite{ren2016faster} for the detection head. Training is conducted for 600 epochs with a batch size of 4, using the Adam optimizer with an initial learning rate of 1e-3, which gradually decays to 1e-7.

Quantitative results are presented in Tab. \ref{tab: object detection}. The proposed method achieves superior performance in most categories. Notably, our method demonstrates a considerable gain in the motorbike class, surpassing the infrared method by more than 6.97. This class is particularly challenging due to its rarity compared to other categories. As shown in Tab. \ref{tab: object detection}, despite the comparison methods employing more powerful backbones, DeFusion++ achieves significantly better results. These outcomes highlight the advantages of the design of our self-supervised learning framework.

\begin{table}
\centering
\caption{Performance comparison of object detection on the M3FD dataset.}
\label{tab: object detection}
\setlength{\tabcolsep}{2.5pt} 
\begin{tabular}{lcccccccc}
\toprule
{Method} & {People} & {Car} & {Bus} & {Lamp} & {Motor} & {Truck} & AP\textsubscript{50} & {AP} \\
\midrule
VIS & 74.81 & 89.46 & 79.98 & \textbf{85.46} & 71.03 & 82.28 & 80.50 & 47.97 \\
IR & 86.31 & 83.4 & 79.31 & 61.44 & \underline{72.37} & 82.07 & 77.49 & 45.87 \\
\color{dt} FusionGAN \cite{ma2019fusiongan} & \color{dt} 86.41 & \color{dt} 88.09 & \color{dt} 80.12 & \color{dt} 79.36 & \color{dt} 70.02 & \color{dt} 83.49 & \color{dt} 81.25 & \color{dt} 47.48 \\
\color{dt} CDDFuse \cite{zhao2023cddfuse} & \color{dt} \underline{88.19} & \color{dt} 88.59 & \color{dt}{82.63} & \color{dt} 81.76 & \color{dt} 68.78 & \color{dt} 83.59 & \color{dt} 82.26 & \color{dt} \underline{50.79} \\
\color{dt} MetaFusion \cite{zhou2023dynamics} & \color{dt} 86.60 & \color{dt} 88.57 & \color{dt} \underline{83.70} & \color{dt} 83.17 & \color{dt} 69.69 & \color{dt} 82.31 & \color{dt} 82.34 & \color{dt} 49.58 \\
\color{dt} TIMFusion \cite{liu2024task} & \color{dt} 82.40 & \color{dt} 89.35 & \color{dt} 80.96 & \color{dt} 85.43 & \color{dt} 64.42 & \color{dt} 80.36 & \color{dt} 80.49 & \color{dt} 48.47 \\
IFCNN \cite{zhang2020ifcnn} & 86.70 & \underline{89.49} & 79.42 & \underline{85.34} & 69.60 & \underline{84.17} & \underline{82.45} & 50.75 \\
PMGI \cite{zhang2020rethinking} & 87.62 & 88.50 & 80.90 & 83.01 & 69.19 & 83.96 & 82.20 & 49.85 \\
U2Fusion \cite{xu2020u2fusion} & 87.94 & 89.0 & 81.39 & 84.42 & 65.88 & 82.19 & 81.80 & 50.05 \\
DeFusion++ & \textbf{88.70} & \textbf{90.13} & \textbf{86.30} & 81.55 & \textbf{79.34} & \textbf{84.57} & \textbf{85.10} & \textbf{51.01} \\
\bottomrule
\end{tabular}
\end{table}

\subsubsection{Semantic Segmentation}

\begin{figure*}[htbp]
    \centering
    \includegraphics[width=18cm]{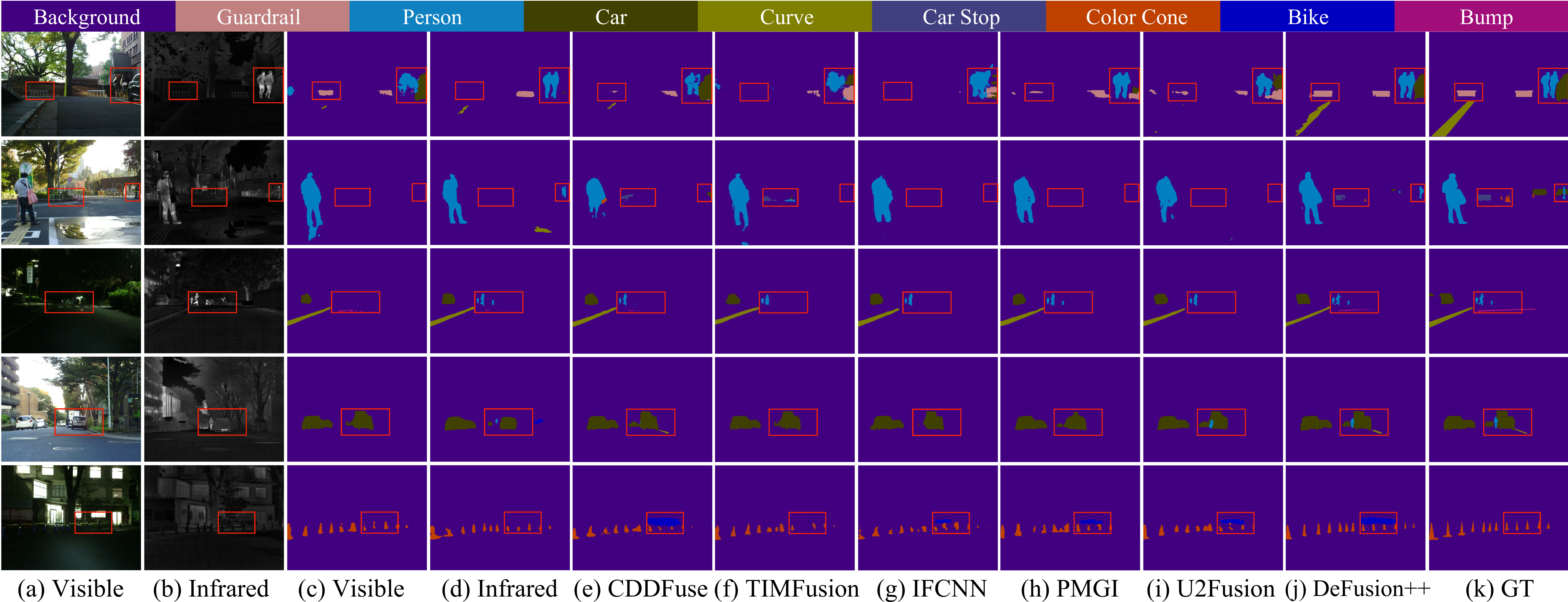}
    \caption{Segmentation results of various MEF fusion algorithms on the MSRS dataset. }
    \label{fig: ivf-seg}
\end{figure*}

For the semantic segmentation task, we utilize the MSRS dataset as our evaluation dataset, which comprises 1,444 pairs of aligned infrared and visible images, categorized into nine high-quality object classes. In dividing the dataset, we follow the guidelines suggested by \cite{tang2022piafusion}. For the segmentation network, we select a robust framework, EVA \cite{fang2023eva}, for comparison with other methods. EVA employs a pretrained vision transformer (ViT-B) as the backbone of the segmentation network. To ensure a fair comparison, both our network and those of the other methods utilize the same segmentation head, namely UPerHead \cite{xiao2018unified}. The quantitative and qualitative results are presented in Tab. \ref{tab: ivf_seg} and Fig. \ref{fig: ivf-seg}, respectively.

The qualitative results across various categories for competitors on the MSRS dataset are presented in Tab. \ref{tab: ivf_seg}. It is evident that DeFusion++ outperforms all other methods across all categories in terms of mean intersection over union (mIOU) and mean accuracy (mAcc). Notably, the proposed method achieves a significant improvement, with a gain of over 7.93 in mIOU compared to IFCNN. Benefiting from the self-supervised training approach, our method is particularly sensitive to rare sample categories, such as the cone. Compared to its counterparts, DeFusion++ demonstrates a significant advantage. These results illustrate that our method is markedly superior to comparison methods, even with fewer learnable parameters.

Five examples of segmentation results on the MSRS test dataset are illustrated in Fig. \ref{fig: ivf-seg}. In the first example, the guardrail is easily recognized in the visible image due to its rich texture, while people are more readily detected in the infrared image, thanks to heat radiation. The segmentation results from both the visible and infrared images corroborate this observation: the visible segmentation results inaccurately depict people, and the infrared segmentation misses the guardrail. All comparison fusion methods fail to clearly segment both elements into a coherent map. Conversely, our method successfully segments both objects, producing segmentation results with clear edges. This indicates that DeFusion++ retains more useful information from the source images. Notably, the last example presents an interesting phenomenon where the bike is difficult to recognize in both infrared and visible images, leading to its omission from the ground truth. However, all fusion methods, except for TIMFusion, identify the object, suggesting that the fused images can to some extent discard noise information. In this comparison, our method delivers the most precise predictions, as the segmentation map includes most of bikes. This demonstrates that our method generalizes well rather than overfitting to the dataset.

\begin{table*}
\caption{Performance comparison of semantic segmentation on the MSRS dataset. Best values are in \textbf{bold} and second-best values are \underline{underlined}.}
\label{tab: ivf_seg}
\setlength{\tabcolsep}{2.2pt} 
\begin{tabular}{lcccccccccccccccccccc}
\toprule
\multirow{2}{*}{Method} & \multicolumn{2}{c}{Unlabel} & \multicolumn{2}{c}{Car} & \multicolumn{2}{c}{Person} & \multicolumn{2}{c}{Bike} & \multicolumn{2}{c}{Curve} & \multicolumn{2}{c}{Car Stop} & \multicolumn{2}{c}{Guail} & \multicolumn{2}{c}{Cone} & \multicolumn{2}{c}{Bump} & \multirow{2}{*}{mIOU} & \multirow{2}{*}{mAcc} \\ \cmidrule(lr){2-3} \cmidrule(lr){4-5} \cmidrule(lr){6-7} \cmidrule(lr){8-9} \cmidrule(lr){10-11} \cmidrule(lr){12-13} \cmidrule(lr){14-15} \cmidrule(lr){16-17} \cmidrule(lr){18-19}
& IoU & Acc & IoU & Acc & IoU & Acc & IoU & Acc & IoU & Acc & IoU & Acc & IoU & Acc & IoU & Acc & IoU & Acc & & \\ 
\midrule
IR & 97.51 & 99.13 & 79.66 & 84.91 & \underline{65.98} & \underline{81.46} & 56.72 & 64.89 & 50.84 & 61.13 & 63.32 & 75.85 & 63.85 & 68.66 & 46.45 & 55.13 & 60.68 & 68.18 & 65.00 & 73.26 \\
Vis & 97.66 & 99.16 & 82.73 & \underline{90.68} & 46.99 & 59.14 & 59.46 & 70.38 & 54.74 & 64.56 & 68.00 & 77.89 & {76.19} & \underline{91.47} & 51.74 & 66.90 & \underline{76.08} & \underline{79.24} & 68.18 & 77.71 \\
\color{dt} FusionGAN \cite{ma2019fusiongan} & \color{dt} 97.47 & \color{dt} 99.18 & \color{dt} 79.57 & \color{dt} 85.59 &\color{dt} 65.43 & \color{dt}77.40 & \color{dt}55.17 & \color{dt}62.73 & \color{dt}53.41 & \color{dt}64.10 & \color{dt}62.27 & \color{dt}75.14 & \color{dt}62.21 & \color{dt}71.68 & \color{dt}49.81 & \color{dt}60.51 & \color{dt}44.15 & \color{dt}45.80 & \color{dt}63.27 & \color{dt}71.35 \\
\color{dt} CDDFuse \cite{zhao2023cddfuse} & \color{dt}97.80 & \color{dt}99.06 & \color{dt}82.70 & \color{dt}\underline{90.68} & \color{dt}{60.86} & \color{dt}{74.31} & \color{dt}59.70 & \color{dt}{70.47} & \color{dt}57.36 & \color{dt}67.53 & \color{dt}68.84 & \color{dt}79.37 & \color{dt}{73.07} & \color{dt}81.39 & \color{dt}53.84 & \color{dt}66.60 & \color{dt}{70.38} & \color{dt}{75.19} & \color{dt}69.40 & \color{dt}78.29 \\
\color{dt} MetaFusion \cite{zhou2023dynamics} & \color{dt}\underline{97.89} & \color{dt}99.20 & \color{dt}82.93 & \color{dt}{90.33} & \color{dt}{56.50} & \color{dt}{68.48} & \color{dt}\underline{62.22} & \color{dt}{70.69} & \color{dt}57.06 & \color{dt}65.99 & \color{dt}\underline{70.63} & \color{dt}\underline{81.07} & \color{dt}\underline{79.25} & \color{dt}85.68 & \color{dt}52.36 & \color{dt}63.06 & \color{dt}{70.67} & \color{dt}{77.71} & \color{dt}69.95 & \color{dt}78.02 \\
\color{dt} TIMFusion \cite{liu2024task} & \color{dt}97.62 & \color{dt}99.16 & \color{dt}82.47 & \color{dt}{89.81} & \color{dt}{50.48} & \color{dt}{59.90} & \color{dt}60.46 & \color{dt}\underline{70.76} & \color{dt}46.12 & \color{dt}54.51 & \color{dt}66.09 & \color{dt}75.64 & \color{dt}{77.11} & \color{dt}88.58 & \color{dt}51.23 & \color{dt}62.70 & \color{dt}{70.50} & \color{dt}{76.17} & \color{dt}66.90 & \color{dt}75.25 \\
IFCNN \cite{zhang2020ifcnn} & {97.88} & {99.19} & \underline{83.22} & 89.52 & 61.79 & 75.39 & {60.79} & 69.16 & \underline{57.71} & \underline{67.58} & {70.04} & {80.22} & 75.34 & 88.18 & \underline{54.88} & \underline{67.09} & 69.09 & 73.45 & \underline{70.08} & \underline{78.86} \\
PMGI \cite{zhang2020rethinking} & 97.76 & 99.22 & 82.29 & 88.25 & 61.71 & 73.43 & 57.91 & 68.13 & 55.76 & 65.19 & 60.00 & 69.16 & 62.96 & 72.27 & 51.67 & 63.33 & 68.75 & 73.84 & 66.53 & 74.76 \\
U2Fusion \cite{xu2020u2fusion} & 97.73 & \underline{99.23} & 82.32 & 88.45 & 59.55 & 70.87 & 58.63 & {68.18} & 54.65 & 63.83 & 63.22 & 72.60 & 64.39 & 75.65 & 51.36 & 60.85 & 66.51 & 71.80 & 66.48 & 74.61 \\
DeFusion++ & \textbf{98.47} & \textbf{99.28} & \textbf{87.91} & \textbf{93.49} & \textbf{73.68} & \textbf{85.93} & \textbf{69.74} & \textbf{80.60} & \textbf{65.18} & \textbf{76.27} & \textbf{78.16} & \textbf{87.21} & \textbf{85.14} & \textbf{91.75} & \textbf{62.51} & \textbf{72.79} & \textbf{81.30} & \textbf{86.56} & \textbf{78.01} & \textbf{85.99} \\
\bottomrule
\end{tabular}
\end{table*}

\subsection{Ablation Study}
\subsubsection{The Effectiveness of MCUD}

\begin{figure}[htbp]
    \centering
    \begin{overpic}[width=9cm]{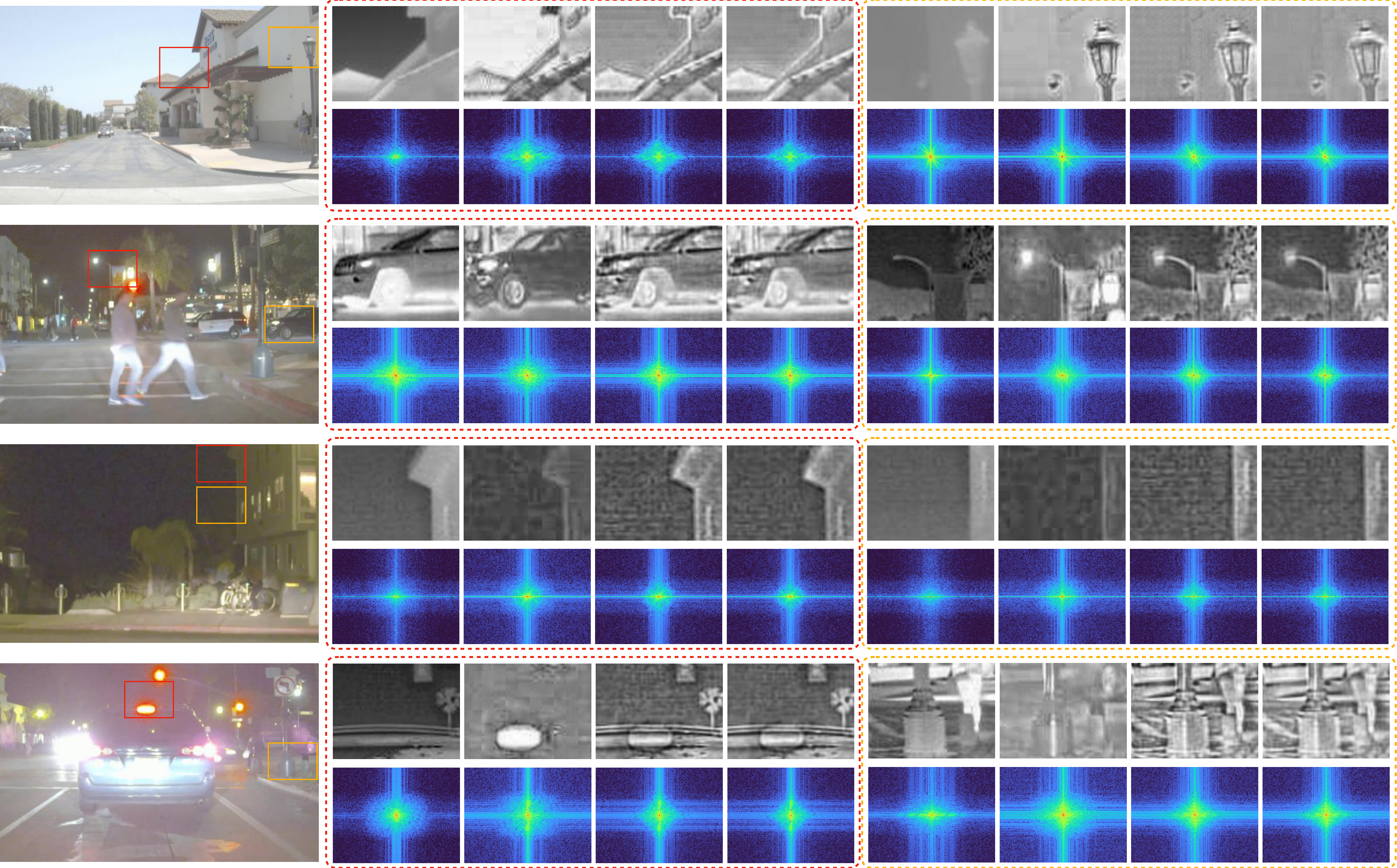}
    \put(27.0, -2.7){\color{black}\scriptsize  Inf.}
        \put(35.7, -2.7){\color{black}\scriptsize  Vis.}
        \put(44.0, -2.7){\color{black}\scriptsize  CUD}
        \put(52.3, -2.7){\color{black}\scriptsize  MCUD}
        \put(65.4, -2.7){\color{black}\scriptsize  Inf.}
        \put(74.3, -2.7){\color{black}\scriptsize  Vis.}
        \put(82.0, -2.7){\color{black}\scriptsize  CUD}
        \put(90.2, -2.7){\color{black}\scriptsize  MCUD}
    \end{overpic}
    \caption{Qualitative results of the effectiveness of MUCD.}
    \label{fig: mcud-fused}
\end{figure}

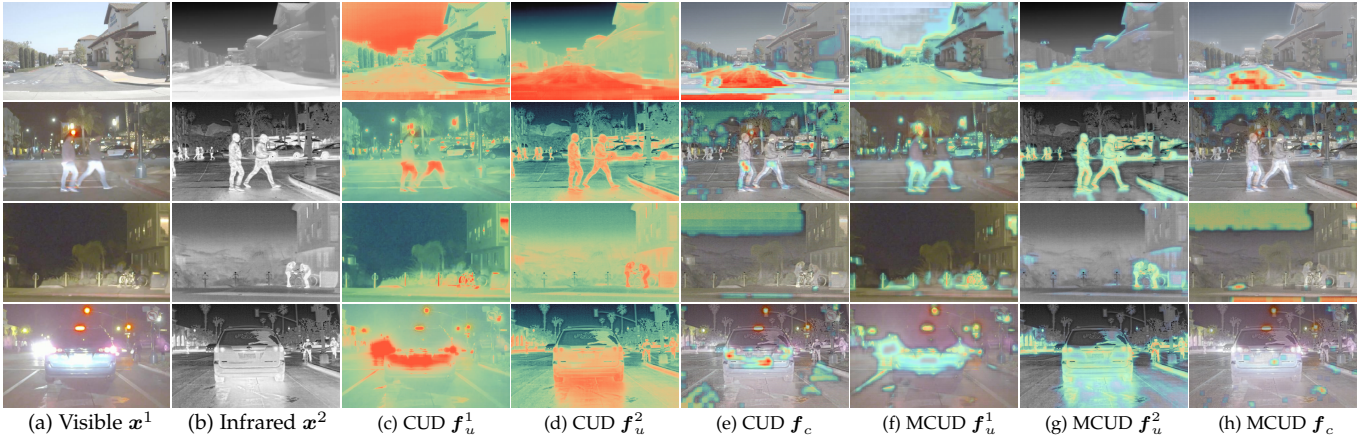
\begin{figure*}[htbp]
    \centering
    \begin{overpic}[width=18cm]{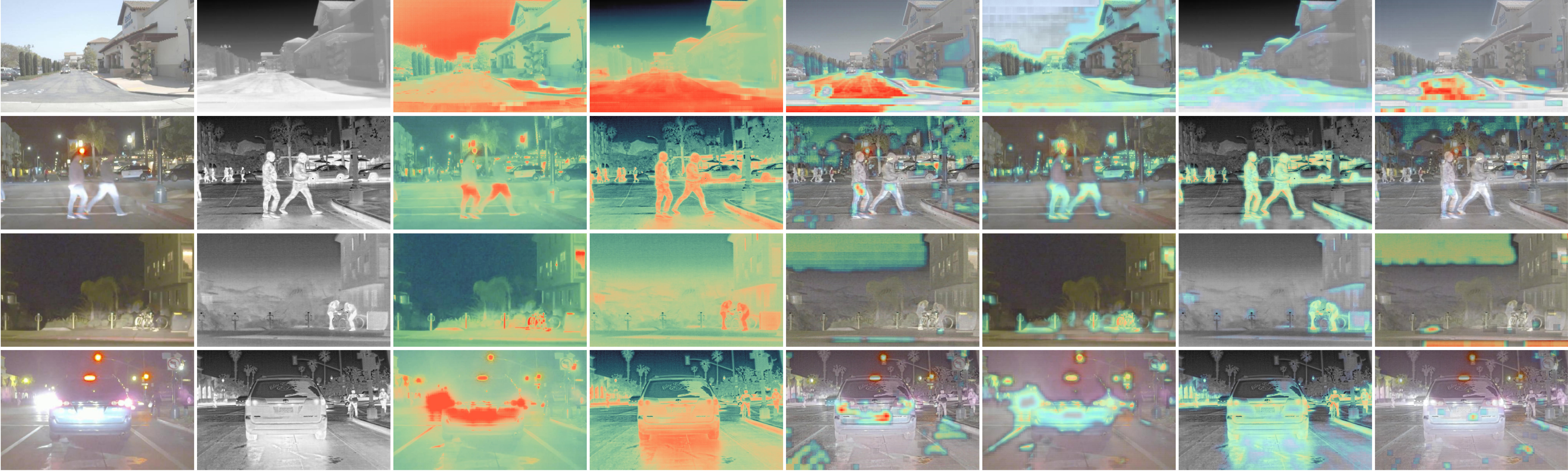}
    \put(1.9, -1.5){\color{black}\footnotesize (a) Visible $\ebf{x}^1$}
        \put(13.7, -1.5){\color{black}\footnotesize (b) Infrared $\ebf{x}^2$}
        \put(27.7, -1.5){\color{black}\scriptsize (c) CUD $\ebf{f}_u^1$}
        \put(40.0, -1.5){\color{black}\scriptsize (d) CUD $\ebf{f}_u^2$}
        \put(52.7, -1.5){\color{black}\scriptsize (e) CUD $\ebf{f}_c$}
        \put(65.0, -1.5){\color{black}\scriptsize (f) MCUD $\ebf{f}_u^1$}
        \put(77.0, -1.5){\color{black}\scriptsize (g) MCUD $\ebf{f}_u^2$}
        \put(89.8, -1.5){\color{black}\scriptsize (h) MCUD $\ebf{f}_c$}
    \end{overpic}
    \caption{Qualitative results of the effectiveness of MUCD. To enhance visual clarity, feature maps are overlaid onto the source images. In the visualized feature maps, color intensity represents feature values, with deeper red indicating higher values.}
    \label{fig: mcud-decom}
\end{figure*}

In DeFusion++, we design a multi-modal common and unique decomposition (MCUD) for the multi-modal fusion task. To validate its effectiveness, we conduct a comparative experiment where tasks are trained solely using CUD. We analyze the effectiveness from two aspects: (i) the quality of the fusion results, and (ii) the visualization of the common and unique representations. The results are presented in Figs. \ref{fig: mcud-fused} and \ref{fig: mcud-decom}, respectively.

A comparison of fused images from the RoadScene dataset is presented in Fig. \ref{fig: mcud-fused}. In this figure, we display cropped patches alongside their corresponding frequency maps. For improved visualization, we convert the images into YCbCr mode and specifically show the Y-channel images. It is evident that models not employing multi-modal CUD tend to retain more noise in the fused results, as illustrated in the third example in the background regions, and exhibit edge artifacts, highlighted by the red box in the first example. The frequency map further reveals these observations. By incorporating multi-modal CUD, the model preserves essential frequency information in the fused results while eliminating unimportant high-frequency noise and perturbations. Consequently, the fused results generated from the model equipped with multi-modal CUD are sparser than those from the model trained only with CUD, especially in high-frequency regions.

The visualization of decomposition components on the RoadScene dataset is depicted in Fig. \ref{fig: mcud-decom}. In this figure, we extract both the common and unique representations from the pretrained model and apply these features to the source images. As demonstrated, when the model is trained without multi-modal CUD, it tends to classify modal-related information predominantly as unique components. As a result, it classifies the entire image as unique information, which contradicts the objectives of CUD tasks. Conversely, the model trained with multi-modal CUD exhibits a robust ability to decompose effectively. Specifically, the model recognizes that the intensity of pixels in infrared images is more useful, while the light in visible images may cause over-exposure, thus rendering some semantic information useless, as illustrated in the last example. The visualization of decomposition demonstrates the effectiveness of MCUD, which facilitates the model in learning more accurate unique and common information.

In the quantitative comparison, we employ a visual strategy to showcase the decomposition components. When comparing the visual results, the model equipped with the parameter-shared cross attention layer demonstrates a performance advantage. The most notable advantage is that the parameter-shared cross attention layer produces more accurate unique decompositions. For instance, the first example illustrates shortcomings where the non-parameter-shared cross attention layer fails to highlight the unique information of the under-exposed image. Similarly, in the last example, the model struggles to accurately represent unique information in the over-exposed image, where the most unique regions are not allocated the greatest attention. In contrast, the parameter-shared cross attention layer accurately captures and represents the most unique information in both over- and under-exposed images. Another advantage of this design is that it reduces the number of learnable parameters. The visualization demonstrates that the parameter-shared cross attention layer effectively exchanges information from different source images.

\subsubsection{The Effectiveness of MFM}

The objective of masked feature modeling (MFM) is to enhance the ability of learning representations. In this ablation study, we verify its effectiveness by conducting comparative experiments on downstream tasks, such as object detection. The results are reported in Tab. \ref{tab: abla obd}. By incorporating MFM training, we observe that the model achieves competitive improvement. Specifically, in the motorbike category, the model trained with MFM achieves a gain of 10 in terms of IOU metrics compared to the model not trained with MFM. This demonstrates that the tailored design of this self-supervised learning task facilitates the model in producing stronger representations.

\begin{table}
\centering
\caption{Ablation study on the impact of the MFM module for object detection utilizing the M3FD dataset.}
\label{tab: abla obd}
\setlength{\tabcolsep}{2.5pt} 
\begin{tabular}{lcccccccc}
\toprule
{Method} & {People} & {Car} & {Bus} & {Lamp} & {Motor} & {Truck} & AP\textsubscript{50} & {AP} \\
\midrule

w/o MFM & {88.08} & {89.86} & {85.38} & \textbf{83.45} & {69.14} & \textbf{87.91} & {83.97} & {49.32} \\
w MFM& \textbf{88.70} & \textbf{90.13} & \textbf{86.30} & {81.55} & \textbf{79.34} & {84.57} & \textbf{85.10} & \textbf{51.01} \\
\bottomrule
\end{tabular}
\end{table}

\subsubsection{DeFusion++ versus DeFusion}

\begin{figure*}[htbp]
    \centering
     \begin{overpic}[width=18cm]{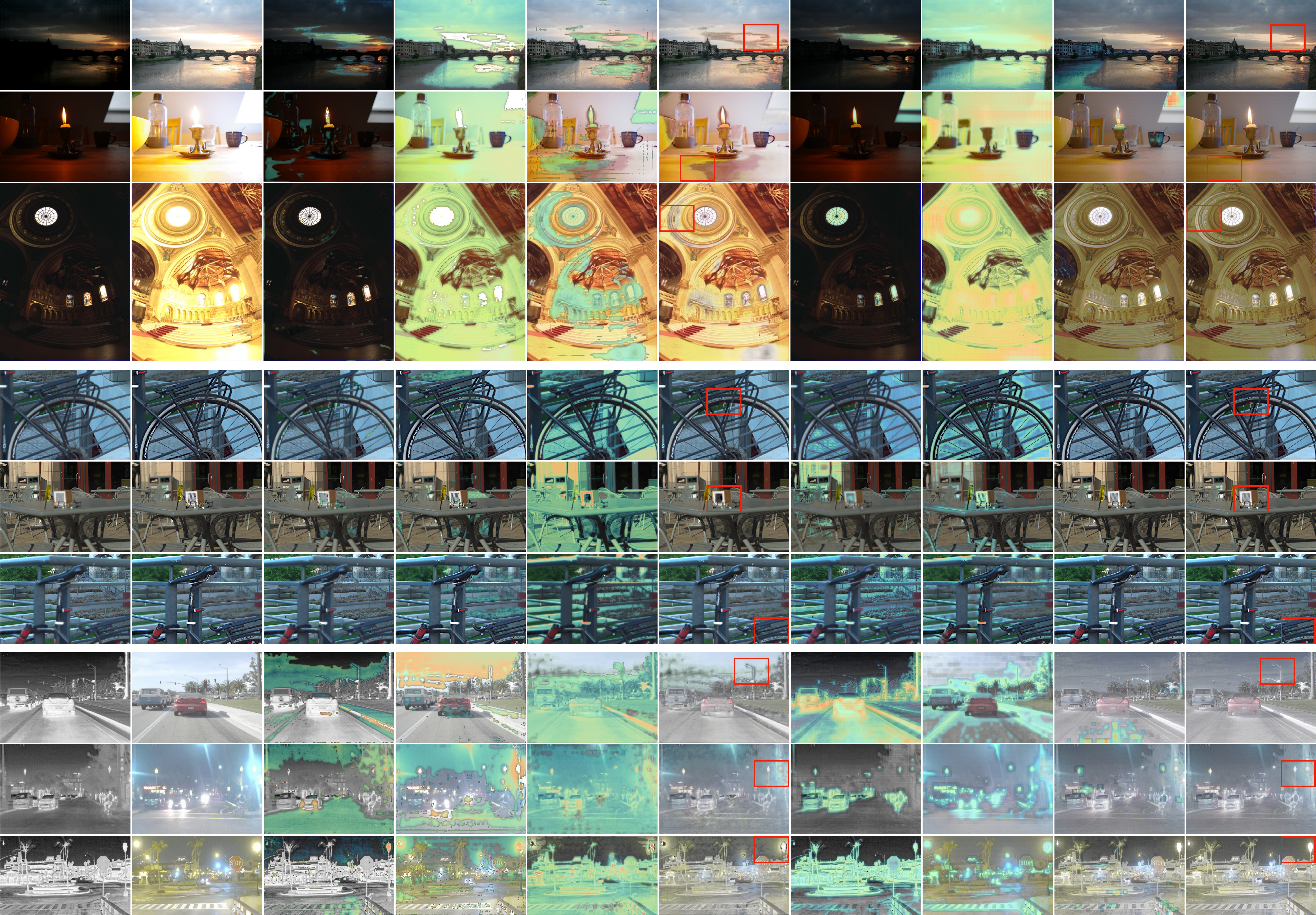}
    \put(1, -1.5){\color{black}\scriptsize (a) Image 1 $\ebf{x}^1$}
        \put(11.0, -1.5){\color{black}\scriptsize (b) Image 2 $\ebf{x}^2$}
        \put(20.0, -1.5){\color{black}\scriptsize (c) DeFusion $\ebf{f}_u^1$}
        \put(30.0, -1.5){\color{black}\scriptsize (d) DeFusion $\ebf{f}_u^2$}
        \put(40.5, -1.5){\color{black}\scriptsize (e) DeFusion $\ebf{f}_c$}
        \put(51.7, -1.5){\color{black}\scriptsize (f) Fusion}
        \put(60.0, -1.5){\color{black}\scriptsize (g) \tiny DeFusion++ \scriptsize $\ebf{f}_u^1$}
        \put(70.0, -1.5){\color{black}\scriptsize (h) \tiny DeFusion++ \scriptsize $\ebf{f}_u^2$}
        \put(81.2, -1.5){\color{black}\scriptsize (i) \tiny DeFusion++ \scriptsize $\ebf{f}_c$}
        \put(91.8, -1.5){\color{black}\scriptsize (j) Fusion}
    \end{overpic}
    \caption{Comparison of our DeFusion++ with the original DeFusion. }
    \label{fig: different DeFusion}
\end{figure*}

The preliminary version of the proposed method and its enhancements are detailed in Sec. \ref{sec:introduction}. To validate the effectiveness of these improvements, we compare the results of DeFusion++ with those of previous version of DeFusion, as illustrated in Fig. \ref{fig: different DeFusion}. To enhance the visualization, we map the decomposition components onto a heatmap and then overlay this heatmap onto the source image.

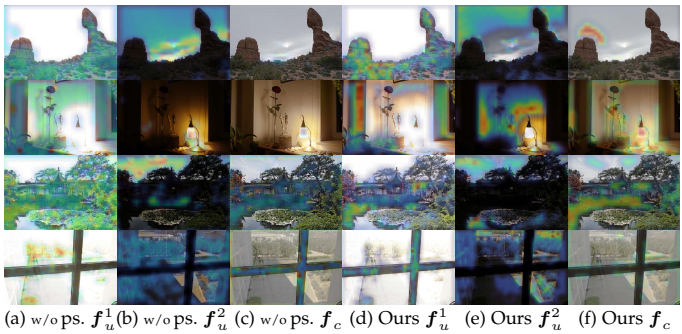
\begin{figure}[htbp]
    \centering
    \begin{overpic}[width=9cm]{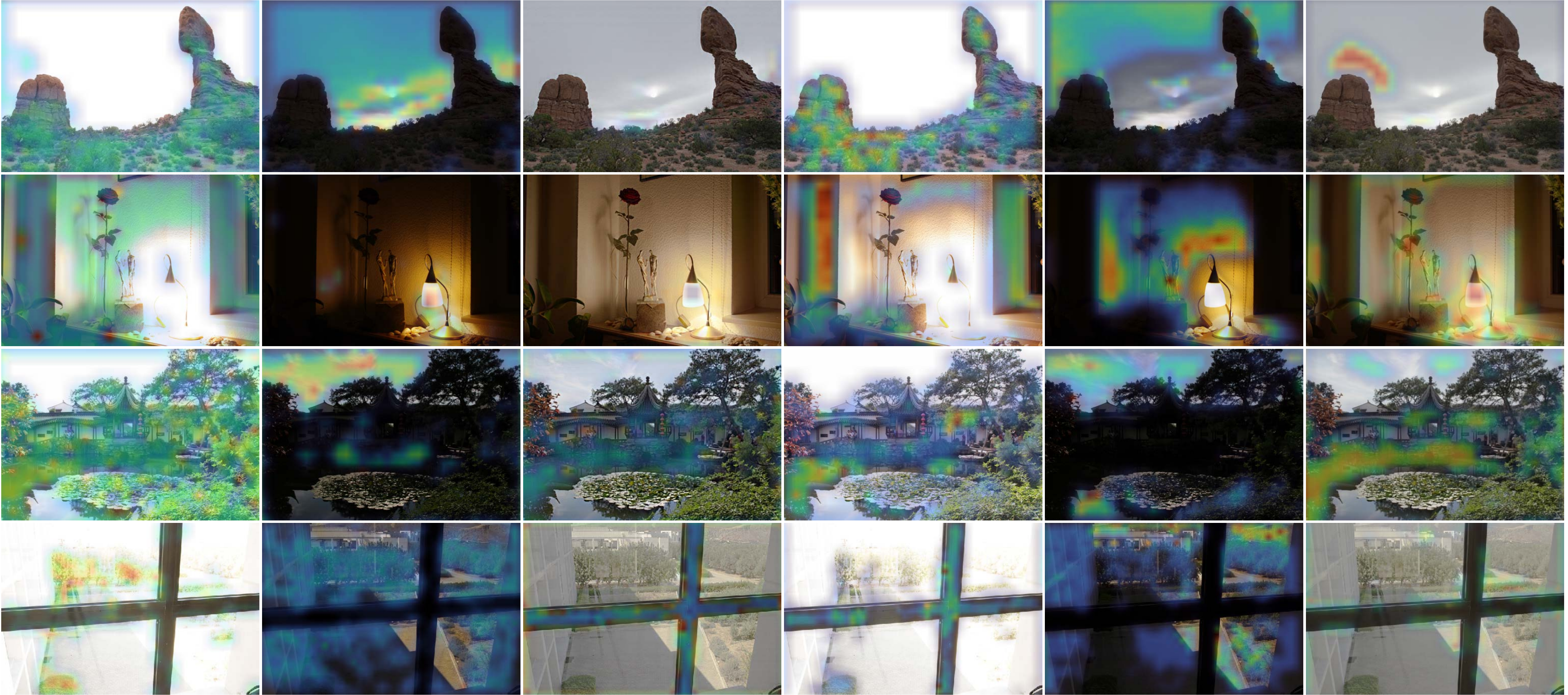}
        \put(0, -2.9){\color{black}\scriptsize (a) \tiny w/o \scriptsize ps. $\ebf{f}_u^1$}
        \put(16.5, -2.9){\color{black}\scriptsize (b) \tiny w/o \scriptsize ps. $\ebf{f}_u^2$}
        \put(34.0, -2.9){\color{black}\scriptsize (c) \tiny w/o \scriptsize ps. $\ebf{f}_c$}
        \put(51.0, -2.9){\color{black}\scriptsize (d) Ours $\ebf{f}_u^1$}
        \put(68.0, -2.9){\color{black}\scriptsize (e) Ours $\ebf{f}_u^2$}
        \put(84.8, -2.9){\color{black}\scriptsize (f) Ours $\ebf{f}_c$}
    \end{overpic}
    \caption{Qualitative results of the cross attention layer. }
    \label{fig: cross-attention}
\end{figure}

Firstly, we enhance the backbone of the fusion network by incorporating a vision transformer equipped with global attention, which facilitates the capture of fine, holistic semantic information. The impact of this improvement is highlighted by the red boxes in the first three examples in Fig. \ref{fig: different DeFusion}. In DeFusion, the decomposition components are inconsistent and inaccurate, seriously affecting the common representation and resulting in artifacts within the fused results. In contrast, DeFusion++ features unique and common representations that are essentially complementary, aiding the network in producing visually pleasing fused results. In the multi-focus examples, namely the fourth, fifth, and sixth examples in Fig. \ref{fig: different DeFusion}, DeFusion++ exhibits superior representational ability. It accurately identifies foreground and background elements, focusing more clearly on distinct regions even in complex, spatially varying scenes, as demonstrated in the fourth example.

Secondly, we have modified the CUD task to accommodate multi-modal image fusion. The effectiveness of this design is illustrated in the last three examples of Fig. \ref{fig: different DeFusion}. The decomposed unique representations of DeFusion incorporates substantial modality-related information. For instance, DeFusion tends to focus excessively on the sky in infrared images, where the sky appears black. However, in infrared imagery, pixel intensity is the most critical information. From this perspective, DeFusion fails to extract the most effective representations from the source image. In contrast, DeFusion++ is trained on multi-modal images, which facilitates a deeper semantic understanding of the source images. As a result, DeFusion++ effectively highlights salient objects in the infrared image, typically characterized by higher pixel values, while avoiding the preservation of over-exposed regions that also exhibit high pixel values in visible images, as demonstrated in the eighth example.

Lastly, DeFusion++ naturally extends to downstream high-level tasks. In this paper, we verify that DeFusion++ achieves SOTA performance in multiple downstream tasks beyond just image fusion. In contrast, DeFusion was designed as a general image fusion network, which does not provide representations that can be directly adapted to downstream tasks.

In DeFusion++, we utilize a parameter-shared cross-attention layer to directly extract common and unique features. Specifically, during the generation of unique features, this layer captures modality-related unique features by processing different tokens. To evaluate the effectiveness of the parameter-shared design, we visualize the decomposition features and compare the fused results. Additionally, we design a non-parameter-shared cross-attention layer to serve as an ablation control. We conduct this study on the multi-exposure task using the MEFB dataset. The quantitative comparison is shown in Fig. \ref{fig: cross-attention}, we employ a visual strategy to showcase the decomposition components. When comparing the visual results, the model equipped with the parameter-shared cross attention layer demonstrates a performance advantage. The most notable advantage is that the parameter-shared cross attention layer produces more accurate unique decompositions. For instance, the first example illustrates shortcomings where the non-parameter-shared cross attention layer fails to highlight the unique information of the under-exposed image. Similarly, in the last example, the model struggles to accurately represent unique information in the over-exposed image, where the most unique regions are not allocated the greatest attention. In contrast, the parameter-shared cross attention layer accurately captures and represents the most unique information in both over- and under-exposed images. Another advantage of this design is that it reduces the number of learnable parameters. The visualization demonstrates that the parameter-shared cross attention layer effectively exchanges information from different source images.

\section{Conclusion}

In this study, we introduce DeFusion++, a novel self-supervised image fusion framework that redefines traditional image fusion approaches. DeFusion++ strategically decomposes images into unique and common components, streamlining the fusion process and enhancing output utility and robustness through two innovative pretext tasks: common and unique decomposition (CUD) and masked feature modeling (MFM). Through rigorous evaluations on diverse tasks such as infrared-visible fusion, multi-focus fusion, and multi-exposure fusion, DeFusion++ not only improves the aesthetic quality of fused images but also their effectiveness in critical downstream applications like image segmentation and object detection. Performance assessments on three datasets demonstrate that DeFusion++ surpasses current methods in adaptability and efficiency. The ability of DeFusion++ to produce adaptable fused representations showcases its potential as a foundational technology for applications requiring accurate and comprehensive visual information. DeFusion++ is poised to drive innovation in image processing and artificial intelligence, setting a new benchmark in the field and broadening the scope of image fusion applications.





\ifCLASSOPTIONcaptionsoff
  \newpage
\fi



%

{
\bibliographystyle{IEEEtran}
\bibliography{referencev2}

\begin{thebibliography}{10}
\providecommand{\url}[1]{#1}
\csname url@samestyle\endcsname
\providecommand{\newblock}{\relax}
\providecommand{\bibinfo}[2]{#2}
\providecommand{\BIBentrySTDinterwordspacing}{\spaceskip=0pt\relax}
\providecommand{\BIBentryALTinterwordstretchfactor}{4}
\providecommand{\BIBentryALTinterwordspacing}{\spaceskip=\fontdimen2\font plus
\BIBentryALTinterwordstretchfactor\fontdimen3\font minus \fontdimen4\font\relax}
\providecommand{\BIBforeignlanguage}[2]{{%
\expandafter\ifx\csname l@#1\endcsname\relax
\typeout{** WARNING: IEEEtran.bst: No hyphenation pattern has been}%
\typeout{** loaded for the language `#1'. Using the pattern for}%
\typeout{** the default language instead.}%
\else
\language=\csname l@#1\endcsname
\fi
#2}}
\providecommand{\BIBdecl}{\relax}
\BIBdecl

\bibitem{zhang2021image}
H.~Zhang, H.~Xu, X.~Tian, J.~Jiang, and J.~Ma, ``Image fusion meets deep learning: A survey and perspective,'' \emph{Information Fusion}, vol.~76, pp. 323--336, 2021.

\bibitem{ma2024reciprocal}
Q.~Ma, J.~Jiang, X.~Liu, and J.~Ma, ``Reciprocal transformer for hyperspectral and multispectral image fusion,'' \emph{Information Fusion}, vol. 104, p. 102148, 2024.

\bibitem{xu2021emfusion}
H.~Xu and J.~Ma, ``{EMFusion}: An unsupervised enhanced medical image fusion network,'' \emph{Information Fusion}, vol.~76, pp. 177--186, 2021.

\bibitem{liu2011objective}
Z.~Liu, E.~Blasch, Z.~Xue, J.~Zhao, R.~Laganiere, and W.~Wu, ``Objective assessment of multiresolution image fusion algorithms for context enhancement in night vision: a comparative study,'' \emph{IEEE Transactions on Pattern Analysis and Mahine Intelligence}, vol.~34, no.~1, pp. 94--109, 2011.

\bibitem{wang2021deep}
L.~Wang and K.-J. Yoon, ``Deep learning for hdr imaging: State-of-the-art and future trends,'' \emph{IEEE Transactions on Pattern Analysis and Mahine Intelligence}, vol.~44, no.~12, pp. 8874--8895, 2021.

\bibitem{ma2019infrared}
J.~Ma, Y.~Ma, and C.~Li, ``Infrared and visible image fusion methods and applications: A survey,'' \emph{Information Fusion}, vol.~45, pp. 153--178, 2019.

\bibitem{ma2020infrared}
J.~Ma, P.~Liang, W.~Yu, C.~Chen, X.~Guo, J.~Wu, and J.~Jiang, ``Infrared and visible image fusion via detail preserving adversarial learning,'' \emph{Information Fusion}, vol.~54, pp. 85--98, 2020.

\bibitem{xu2020u2fusion}
H.~Xu, J.~Ma, J.~Jiang, X.~Guo, and H.~Ling, ``{U2Fusion}: A unified unsupervised image fusion network,'' \emph{IEEE Transactions on Pattern Analysis and Machine Intelligence}, vol.~44, no.~1, pp. 502--518, 2022.

\bibitem{ma2022swinfusion}
J.~Ma, L.~Tang, F.~Fan, J.~Huang, X.~Mei, and Y.~Ma, ``{SwinFusion}: Cross-domain long-range learning for general image fusion via swin transformer,'' \emph{IEEE/CAA Journal of Automatica Sinica}, vol.~9, no.~7, pp. 1200--1217, 2022.

\bibitem{li2018densefuse}
H.~Li and X.-J. Wu, ``{DenseFuse}: A fusion approach to infrared and visible images,'' \emph{IEEE Transactions on Image Processing}, vol.~28, no.~5, pp. 2614--2623, 2018.

\bibitem{liu2017multi}
Y.~Liu, X.~Chen, H.~Peng, and Z.~Wang, ``Multi-focus image fusion with a deep convolutional neural network,'' \emph{Information Fusion}, vol.~36, pp. 191--207, 2017.

\bibitem{ram2017deepfuse}
K.~Ram~Prabhakar, V.~Sai~Srikar, and R.~Venkatesh~Babu, ``{DeepFuse}: A deep unsupervised approach for exposure fusion with extreme exposure image pairs,'' in \emph{Proceedings of the IEEE International Conference on Computer Vision}, 2017, pp. 4714--4722.

\bibitem{liu2024coconet}
J.~Liu, R.~Lin, G.~Wu, R.~Liu, Z.~Luo, and X.~Fan, ``{CoCoNet}: Coupled contrastive learning network with multi-level feature ensemble for multi-modality image fusion,'' \emph{International Journal of Computer Vision}, vol. 132, no.~5, pp. 1748--1775, 2024.

\bibitem{ma2019fusiongan}
J.~Ma, W.~Yu, P.~Liang, C.~Li, and J.~Jiang, ``{FusionGAN}: A generative adversarial network for infrared and visible image fusion,'' \emph{Information Fusion}, vol.~48, pp. 11--26, 2019.

\bibitem{zhang2020rethinking}
H.~Zhang, H.~Xu, Y.~Xiao, X.~Guo, and J.~Ma, ``Rethinking the image fusion: A fast unified image fusion network based on proportional maintenance of gradient and intensity,'' in \emph{Proceedings of the AAAI Conference on Artificial Intelligence}, vol.~34, 2020, pp. 12\,797--12\,804.

\bibitem{zhao2023cddfuse}
Z.~Zhao, H.~Bai, J.~Zhang, Y.~Zhang, S.~Xu, Z.~Lin, R.~Timofte, and L.~Van~Gool, ``{CDDFuse}: Correlation-driven dual-branch feature decomposition for multi-modality image fusion,'' in \emph{Proceedings of the IEEE/CVF Conference on Computer Vision and Pattern Recognition}, 2023, pp. 5906--5916.

\bibitem{liu2022target}
J.~Liu, X.~Fan, Z.~Huang, G.~Wu, R.~Liu, W.~Zhong, and Z.~Luo, ``Target-aware dual adversarial learning and a multi-scenario multi-modality benchmark to fuse infrared and visible for object detection,'' in \emph{Proceedings of the IEEE/CVF Conference on Computer Vision and Pattern Recognition}, 2022, pp. 5802--5811.

\bibitem{tang2022image}
L.~Tang, J.~Yuan, and J.~Ma, ``Image fusion in the loop of high-level vision tasks: A semantic-aware real-time infrared and visible image fusion network,'' \emph{Information Fusion}, vol.~82, pp. 28--42, 2022.

\bibitem{liang2022fusion}
P.~Liang, J.~Jiang, X.~Liu, and J.~Ma, ``Fusion from decomposition: A self-supervised decomposition approach for image fusion,'' in \emph{European Conference on Computer Vision}, 2022, pp. 719--735.

\bibitem{zhang2020ifcnn}
Y.~Zhang, Y.~Liu, P.~Sun, H.~Yan, X.~Zhao, and L.~Zhang, ``{IFCNN}: A general image fusion framework based on convolutional neural network,'' \emph{Information Fusion}, vol.~54, pp. 99--118, 2020.

\bibitem{tang2022piafusion}
L.~Tang, J.~Yuan, H.~Zhang, X.~Jiang, and J.~Ma, ``{PIAFusion}: A progressive infrared and visible image fusion network based on illumination aware,'' \emph{Information Fusion}, vol.~83, pp. 79--92, 2022.

\bibitem{li2023lrrnet}
H.~Li, T.~Xu, X.-J. Wu, J.~Lu, and J.~Kittler, ``{LRRNet}: A novel representation learning guided fusion network for infrared and visible images,'' \emph{IEEE Transactions on Pattern analysis and Machine Intelligence}, vol.~45, no.~9, pp. 11\,040--11\,052, 2023.

\bibitem{zhu2024taskcustomized}
P.~Zhu, Y.~Sun, B.~Cao, and Q.~Hu, ``Task-customized mixture of adapters for general image fusion,'' in \emph{Proceedings of the IEEE/CVF Conference on Computer Vision and Pattern Recognition}, 2024.

\bibitem{li2020nestfuse}
H.~Li, X.-J. Wu, and T.~Durrani, ``{NestFuse}: An infrared and visible image fusion architecture based on nest connection and spatial/channel attention models,'' \emph{IEEE Transactions on Instrumentation and Measurement}, vol.~69, no.~12, pp. 9645--9656, 2020.

\bibitem{deng2020deep}
X.~Deng and P.~L. Dragotti, ``Deep convolutional neural network for multi-modal image restoration and fusion,'' \emph{IEEE Transactions on Pattern Analysis and Machine Intelligence}, vol.~43, no.~10, pp. 3333--3348, 2020.

\bibitem{hu2023zmff}
X.~Hu, J.~Jiang, X.~Liu, and J.~Ma, ``{ZMFF}: Zero-shot multi-focus image fusion,'' \emph{Information Fusion}, vol.~92, pp. 127--138, 2023.

\bibitem{shang2024holistic}
X.~Shang, G.~Li, Z.~Jiang, S.~Zhang, N.~Ding, and J.~Liu, ``Holistic dynamic frequency transformer for image fusion and exposure correction,'' \emph{Information Fusion}, vol. 102, p. 102073, 2024.

\bibitem{xu2023unsupervised}
H.~Xu, L.~Haochen, and J.~Ma, ``Unsupervised multi-exposure image fusion breaking exposure limits via contrastive learning,'' in \emph{Proceedings of the AAAI Conference on Artificial Intelligence}, vol.~37, no.~3, 2023, pp. 3010--3017.

\bibitem{hong2024merf}
W.~Hong, H.~Zhang, and J.~Ma, ``{MERF}: A practical hdr-like image generator via mutual-guided learning between multi-exposure registration and fusion,'' \emph{IEEE Transactions on Image Processing}, vol.~33, pp. 2361--2376, 2024.

\bibitem{jiang2023meflut}
T.~Jiang, C.~Wang, X.~Li, R.~Li, H.~Fan, and S.~Liu, ``{MEFLUT}: Unsupervised 1d lookup tables for multi-exposure image fusion,'' in \emph{Proceedings of the IEEE/CVF International Conference on Computer Vision}, 2023, pp. 10\,542--10\,551.

\bibitem{zou2024enhancing}
Y.~Zou, X.~Li, Z.~Jiang, and J.~Liu, ``Enhancing neural radiance fields with adaptive multi-exposure fusion: A bilevel optimization approach for novel view synthesis,'' in \emph{Proceedings of the AAAI Conference on Artificial Intelligence}, vol.~38, no.~7, 2024, pp. 7882--7890.

\bibitem{qu2024trans2fuse}
L.~Qu, S.~Liu, M.~Wang, S.~Li, S.~Yin, and Z.~Song, ``{Trans2Fuse}: Empowering image fusion through self-supervised learning and multi-modal transformations via transformer networks,'' \emph{Expert Systems with Applications}, vol. 236, p. 121363, 2024.

\bibitem{qu2022transmef}
L.~Qu, S.~Liu, M.~Wang, and Z.~Song, ``{TransMEF}: A transformer-based multi-exposure image fusion framework using self-supervised multi-task learning,'' in \emph{Proceedings of the AAAI Conference on Artificial Intelligence}, vol.~36, no.~2, 2022, pp. 2126--2134.

\bibitem{tang2023rethinking}
L.~Tang, H.~Zhang, H.~Xu, and J.~Ma, ``Rethinking the necessity of image fusion in high-level vision tasks: A practical infrared and visible image fusion network based on progressive semantic injection and scene fidelity,'' \emph{Information Fusion}, vol.~99, p. 101870, 2023.

\bibitem{liu2021self}
X.~Liu, F.~Zhang, Z.~Hou, L.~Mian, Z.~Wang, J.~Zhang, and J.~Tang, ``Self-supervised learning: Generative or contrastive,'' \emph{IEEE Transactions on Knowledge and Data Engineering}, vol.~35, no.~1, pp. 857--876, 2021.

\bibitem{zhou2023dynamics}
X.~Zhou, X.~Liu, H.~Wang, D.~Zhai, J.~Jiang, and X.~Ji, ``On the dynamics under the unhinged loss and beyond,'' \emph{Journal of Machine Learning Research}, vol.~24, 2023.

\bibitem{doersch2015unsupervised}
C.~Doersch, A.~Gupta, and A.~A. Efros, ``Unsupervised visual representation learning by context prediction,'' in \emph{Proceedings of the IEEE International Conference on Computer Vision}, 2015, pp. 1422--1430.

\bibitem{noroozi2018boosting}
M.~Noroozi, A.~Vinjimoor, P.~Favaro, and H.~Pirsiavash, ``Boosting self-supervised learning via knowledge transfer,'' in \emph{Proceedings of the IEEE Conference on Computer Vision and Pattern Recognition}, 2018, pp. 9359--9367.

\bibitem{komodakis2018unsupervised}
N.~Komodakis and S.~Gidaris, ``Unsupervised representation learning by predicting image rotations,'' in \emph{Proceedings of the International Conference on Learning Representations}, 2018.

\bibitem{noroozi2017representation}
M.~Noroozi, H.~Pirsiavash, and P.~Favaro, ``Representation learning by learning to count,'' in \emph{Proceedings of the IEEE International Conference on Computer Vision}, 2017, pp. 5898--5906.

\bibitem{zhang2016colorful}
R.~Zhang, P.~Isola, and A.~A. Efros, ``Colorful image colorization,'' in \emph{Proceedings of the European Conference on Computer Vision}.\hskip 1em plus 0.5em minus 0.4em\relax Springer, 2016, pp. 649--666.

\bibitem{chen2020simple}
T.~Chen, S.~Kornblith, M.~Norouzi, and G.~Hinton, ``A simple framework for contrastive learning of visual representations,'' in \emph{International Conference on Machine Learning}.\hskip 1em plus 0.5em minus 0.4em\relax PMLR, 2020, pp. 1597--1607.

\bibitem{he2020momentum}
K.~He, H.~Fan, Y.~Wu, S.~Xie, and R.~Girshick, ``Momentum contrast for unsupervised visual representation learning,'' in \emph{Proceedings of the IEEE Conference on Computer Vision and Pattern Recognition}, 2020, pp. 9729--9738.

\bibitem{he2022masked}
K.~He, X.~Chen, S.~Xie, Y.~Li, P.~Doll{\'a}r, and R.~Girshick, ``Masked autoencoders are scalable vision learners,'' in \emph{Proceedings of the IEEE Conference on Computer Vision and Pattern Recognition}, 2022, pp. 16\,000--16\,009.

\bibitem{zhou2021ibot}
J.~Zhou, C.~Wei, H.~Wang, W.~Shen, C.~Xie, A.~Yuille, and T.~Kong, ``{iBOT}: Image bert pre-training with online tokenizer,'' \emph{Proceedings of the International Conference on Learning Representations}, 2022.

\bibitem{fang2023corrupted}
Y.~Fang, L.~Dong, H.~Bao, X.~Wang, and F.~Wei, ``Corrupted image modeling for self-supervised visual pre-training,'' in \emph{International Conference on Learning Representations}, 2023.

\bibitem{ren2023tinymim}
S.~Ren, F.~Wei, Z.~Zhang, and H.~Hu, ``{TinyMIM}: An empirical study of distilling mim pre-trained models,'' in \emph{Proceedings of the IEEE/CVF Conference on Computer Vision and Pattern Recognition}, 2023, pp. 3687--3697.

\bibitem{jing2020self}
L.~Jing and Y.~Tian, ``Self-supervised visual feature learning with deep neural networks: A survey,'' \emph{IEEE Transactions on Pattern analysis and Machine Intelligence}, vol.~43, no.~11, pp. 4037--4058, 2020.

\bibitem{vision_transformer}
A.~Dosovitskiy, L.~Beyer, A.~Kolesnikov, D.~Weissenborn, X.~Zhai, T.~Unterthiner, M.~Dehghani, M.~Minderer, G.~Heigold, S.~Gelly, J.~Uszkoreit, and N.~Houlsby, ``An image is worth 16x16 words: Transformers for image recognition at scale,'' in \emph{International Conference on Learning Representations}, 2021.

\bibitem{vaswani2017attention}
A.~Vaswani, N.~Shazeer, N.~Parmar, J.~Uszkoreit, L.~Jones, A.~N. Gomez, {\L}.~Kaiser, and I.~Polosukhin, ``Attention is all you need,'' vol.~30, 2017.

\bibitem{girshick2014rich}
R.~Girshick, J.~Donahue, T.~Darrell, and J.~Malik, ``Rich feature hierarchies for accurate object detection and semantic segmentation,'' in \emph{Proceedings of the IEEE/CVF Conference on Computer Vision and Pattern Recognition}, 2014, pp. 580--587.

\bibitem{ma2019mefnet}
K.~Ma, Z.~Duanmu, H.~Zhu, Y.~Fang, and Z.~Wang, ``Deep guided learning for fast multi-exposure image fusion,'' \emph{IEEE Transactions on Image Processing}, vol.~29, pp. 2808--2819, 2020.

\bibitem{liu2023holoco}
J.~Liu, G.~Wu, J.~Luan, Z.~Jiang, R.~Liu, and X.~Fan, ``{HoLoCo}: Holistic and local contrastive learning network for multi-exposure image fusion,'' \emph{Information Fusion}, vol.~95, pp. 237--249, 2023.

\bibitem{liu2023embracing}
Z.~Liu, J.~Liu, G.~Wu, Z.~Chen, X.~Fan, and R.~Liu, ``Searching a compact architecture for robust multi-exposure image fusion,'' \emph{IEEE Transactions on Circuits and Systems for Video Technology}, 2024.

\bibitem{wu2024hybrid}
G.~Wu, H.~Fu, J.~Liu, L.~Ma, X.~Fan, and R.~Liu, ``Hybrid-supervised dual-search: Leveraging automatic learning for loss-free multi-exposure image fusion,'' in \emph{Proceedings of the AAAI Conference on Artificial Intelligence}, vol.~38, no.~6, 2024, pp. 5985--5993.

\bibitem{zhang2021benchmarking}
X.~Zhang, ``Benchmarking and comparing multi-exposure image fusion algorithms,'' \emph{Information Fusion}, vol.~74, pp. 111--131, 2021.

\bibitem{cai2018learning}
J.~Cai, S.~Gu, and L.~Zhang, ``Learning a deep single image contrast enhancer from multi-exposure images,'' \emph{IEEE Transactions on Image Processing}, vol.~27, no.~4, pp. 2049--2062, 2018.

\bibitem{zeng2014perceptual}
K.~Zeng, K.~Ma, R.~Hassen, and Z.~Wang, ``Perceptual evaluation of multi-exposure image fusion algorithms,'' in \emph{2014 Sixth International Workshop on Quality of Multimedia Experience (QoMEX)}.\hskip 1em plus 0.5em minus 0.4em\relax IEEE, 2014, pp. 7--12.

\bibitem{wang2008performance}
Q.~Wang, Y.~Shen, and J.~Jin, ``Performance evaluation of image fusion techniques,'' \emph{Image Fusion: Algorithms and Applications}, vol.~19, pp. 469--492, 2008.

\bibitem{ma2015perceptual}
K.~Ma, K.~Zeng, and Z.~Wang, ``Perceptual quality assessment for multi-exposure image fusion,'' \emph{IEEE Transactions on Image Processing}, vol.~24, no.~11, pp. 3345--3356, 2015.

\bibitem{zhang2021mff}
H.~Zhang, Z.~Le, Z.~Shao, H.~Xu, and J.~Ma, ``{MFF-GAN}: An unsupervised generative adversarial network with adaptive and gradient joint constraints for multi-focus image fusion,'' \emph{Information Fusion}, vol.~66, pp. 40--53, 2021.

\bibitem{zhang2021mfif}
X.~Zhang, ``Deep learning-based multi-focus image fusion: A survey and a comparative study,'' \emph{IEEE Transactions on Pattern Analysis and Machine Intelligence}, vol.~44, no.~9, pp. 4819--4838, 2021.

\bibitem{zhang2020real}
J.~Zhang, Q.~Liao, S.~Liu, H.~Ma, W.~Yang, and J.-H. Xue, ``{Real-MFF}: A large realistic multi-focus image dataset with ground truth,'' \emph{Pattern Recognition Letters}, vol. 138, pp. 370--377, 2020.

\bibitem{zhao2023metafusion}
W.~Zhao, S.~Xie, F.~Zhao, Y.~He, and H.~Lu, ``{MetaFusion}: Infrared and visible image fusion via meta-feature embedding from object detection,'' in \emph{Proceedings of the IEEE/CVF Conference on Computer Vision and Pattern Recognition}, 2023, pp. 13\,955--13\,965.

\bibitem{liu2024task}
R.~Liu, Z.~Liu, J.~Liu, X.~Fan, and Z.~Luo, ``A task-guided, implicitly-searched and metainitialized deep model for image fusion,'' \emph{IEEE Transactions on Pattern Analysis and Machine Intelligence}, 2024.

\bibitem{Toet2014}
\BIBentryALTinterwordspacing
A.~Toet, ``{{TNO} Image Fusion Dataset},'' 4 2014. [Online]. Available: \url{https://figshare.com/articles/dataset/TNO_Image_Fusion_Dataset/1008029}
\BIBentrySTDinterwordspacing

\bibitem{wang2023exploring}
J.~Wang, K.~C. Chan, and C.~C. Loy, ``Exploring clip for assessing the look and feel of images,'' in \emph{Proceedings of the AAAI Conference on Artificial Intelligence}, vol.~37, no.~2, 2023, pp. 2555--2563.

\bibitem{li2022exploring}
Y.~Li, H.~Mao, R.~Girshick, and K.~He, ``Exploring plain vision transformer backbones for object detection,'' in \emph{European Conference on Computer Vision}.\hskip 1em plus 0.5em minus 0.4em\relax Springer, 2022, pp. 280--296.

\bibitem{ren2016faster}
S.~Ren, K.~He, R.~Girshick, and J.~Sun, ``{Faster R-CNN}: Towards real-time object detection with region proposal networks,'' \emph{IEEE Transactions on Pattern Analysis and Machine Intelligence}, vol.~39, no.~6, pp. 1137--1149, 2016.

\bibitem{fang2023eva}
Y.~Fang, W.~Wang, B.~Xie, Q.~Sun, L.~Wu, X.~Wang, T.~Huang, X.~Wang, and Y.~Cao, ``{EVA}: Exploring the limits of masked visual representation learning at scale,'' in \emph{Proceedings of the IEEE/CVF Conference on Computer Vision and Pattern Recognition}, 2023, pp. 19\,358--19\,369.

\bibitem{xiao2018unified}
T.~Xiao, Y.~Liu, B.~Zhou, Y.~Jiang, and J.~Sun, ``Unified perceptual parsing for scene understanding,'' in \emph{Proceedings of the European Conference on Computer Vision}, 2018, pp. 418--434.

\end{thebibliography}
}

\end{document}